\documentclass[final]{IEEEtran}

\usepackage[utf8]{inputenc}  
\usepackage[T1]{fontenc}     
\usepackage{hyperref}        
\usepackage{url}             
\usepackage{booktabs}        
\usepackage{multirow}        
\usepackage{amsfonts}        
\usepackage{amssymb}         
\usepackage{amsthm}          
\usepackage{amsmath}         
\usepackage{nicefrac}        
\usepackage{microtype}       
\usepackage[english]{babel}  
\usepackage{color}           
\usepackage{graphicx}        

\definecolor{orange}{rgb}{1, 0.65, 0}

\newcommand{\M}{\mathcal{M}}
\newcommand{\R}{\mathbb{R}}
\newcommand{\X}{\times}
\newcommand{\norm}[2]{\left\Vert #1 \right\Vert_{{#2}}}
\newcommand{\Exp}{\mathrm{Exp}}
\newcommand{\Log}{\mathrm{Log}}
\newcommand{\mc}{\mathcal}
\renewcommand{\vec}[1]{\boldsymbol{\mathtt{#1}}}

\DeclareMathOperator{\arccosh}{arccosh}
\DeclareMathOperator*{\argmin}{arg\,min}

\begin{document}
\title{Change Detection in Graph Streams by Learning Graph Embeddings on Constant-Curvature Manifolds}
\author{Daniele Grattarola,~\IEEEmembership{Student Member,~IEEE},
        Daniele Zambon,~\IEEEmembership{Student Member,~IEEE},
        Cesare Alippi,~\IEEEmembership{Fellow,~IEEE},
        and Lorenzo Livi,~\IEEEmembership{Member,~IEEE}

\thanks{Daniele Grattarola is with the Faculty of Informatics, 
Universit\`{a} della Svizzera italiana, 
Lugano, Switzerland 
(e-mail: daniele.grattarola@usi.ch).}
\thanks{Daniele Zambon is with the Faculty of Informatics, 
Universit\`{a} della Svizzera italiana, 
Lugano, Switzerland 
(e-mail: daniele.zambon@usi.ch).}
\thanks{Cesare Alippi is with the Dept. of Electronics, Information, and Bioengineering, 
Politecnico di Milano, 
Milan, Italy and 
Faculty of Informatics, 
Universit\`{a} della Svizzera italiana, 
Lugano, Switzerland 
(e-mail: cesare.alippi@polimi.it, cesare.alippi@usi.ch)}
\thanks{Lorenzo Livi is with the Departments of Computer Science and Mathematics, 
University of Manitoba, Winnipeg, MB R3T 2N2, Canada and the Department of Computer Science, College of Engineering, Mathematics and Physical Sciences, 
University of Exeter, 
Exeter EX4 4QF, United Kingdom 
(e-mail: lorenz.livi@gmail.com).}
}

\maketitle
\begin{abstract}
The space of graphs is often characterised by a non-trivial geometry, which complicates learning and inference in practical applications. A common approach is to use embedding techniques to represent graphs as points in a conventional Euclidean space, but non-Euclidean spaces have often been shown to be better suited for embedding graphs. Among these, constant-curvature Riemannian manifolds (CCMs) offer embedding spaces suitable for studying the statistical properties of a graph distribution, as they provide ways to easily compute metric geodesic distances. 
In this paper, we focus on the problem of detecting changes in stationarity in a stream of attributed graphs. To this end, we introduce a novel change detection framework based on neural networks and CCMs, that takes into account the non-Euclidean nature of graphs.
Our contribution in this work is twofold. First, via a novel approach based on adversarial learning, we compute graph embeddings by training an autoencoder to represent graphs on CCMs. Second, we introduce two novel change detection tests operating on CCMs.
We perform experiments on synthetic data, as well as two real-world application scenarios: the detection of epileptic seizures using functional connectivity brain networks, and the detection of hostility between two subjects, using human skeletal graphs. 
Results show that the proposed methods are able to detect even small changes in a graph-generating process, consistently outperforming approaches based on Euclidean embeddings.
\end{abstract}

\section{Introduction}
\label{sec:introduction}

Many relevant machine learning applications require to go beyond conventional Euclidean geometry, as in the case of data described by attributed graphs \cite{richiardi2013machine,bronstein2016geometric}. 
When studying problems on graphs, one of the key issues is to find representations that allow dealing with their underlying geometry, which is usually defined by application-specific distances that often do not satisfy the identity of indiscernibles or the triangular inequality \cite{gm_survey,wilson2014spherical}. The use of metric distances, like graph alignment distances \cite{jain2016geometry}, only mitigates the problem, as they are computationally intractable and hence not useful in practical applications.
Therefore, a common approach is to embed graphs on a more conventional geometric space, such as the Euclidean one.
However, Euclidean geometry is not always the optimal choice, as graphs may find a natural representation on non-Euclidean domains \cite{zambon2018anomaly}. 

Representing graphs as points in metric spaces yields significant benefits when dealing with problems that require studying their statistical properties, and several works in the literature propose relevant manifold learning techniques to approximate high-dimensional data spaces with a low-dimensional representation. However, the computational load required to learn a non-Euclidean manifold and to compute geodesic distances between points is non-negligible \cite{lin2008riemannian}. More importantly, at the current level of research, we lack a solid statistical framework to perform inferential analyses on a learned manifold.
Conversely, constant-curvature manifolds (CCMs), like hyperspherical and hyperbolic spaces, provide a versatile family of non-Euclidean geometries that preserve a metric structure and which can be efficiently computed in a closed form and, therefore, are suitable for inference procedures. Moreover, the considered CCMs have the advantage of being parametrised in a single scalar parameter (the curvature), which completely characterises their geometry \cite{wilson2014spherical}.

In many application scenarios, graphs are assumed to be generated by a stationary process, implying that the topology and graph attributes are drawn from a fixed, albeit unknown, distribution \cite{newman2010networks}. However, the stationarity assumption does not always hold true, with relevant examples including cyber-physical systems \cite{alippi2017not}, functional networks associated with brain imaging (where neural activity changes over time autonomously, or by reaction to stimuli) \cite{heitmann2017putting}, and many others, e.g., see \cite{7296710, masuda2016guide,li2017fundamental,li2018dcrnn_traffic}.

In this paper, we focus on the rather unexplored problem of detecting changes in stationarity of a process generating attributed graphs, hence monitoring whether the common assumption of i.i.d.\ observations breaks.
We show that, by representing graphs on CCMs, we obtain a significant improvement in the detection of changes w.r.t. Euclidean representations.
The contribution of this work is twofold.
First, we propose to use a graph autoencoder \cite{simonovsky2018graphvae, kipf2016variational} to embed graphs on a CCM.
To this end, we introduce a novel approach based on the framework of adversarial autoencoders (AAEs) \cite{makhzani2016adversarial} to impose a geometric constraint on an AAE's latent space, by matching the aggregated posterior of the encoder network with a prior distribution defined on a CCM. By enforcing a prior with support on the CCM, we are able to implicitly impose the geometric constraint, as the AAE will learn to embed graphs on the CCM in order to fool the discriminator network. 
We also propose an AAE that operates without a prior distribution, enforcing the geometric constraint explicitly through a parameter-free discriminator, thus significantly reducing the overall c complexity.
In addition to hyperspherical and hyperbolic latent spaces, we also propose to use an ensemble of different geometries learned by optimising the network to represent the data on several CCMs at the same time. To the best of our knowledge, the presented work is the first to deal with graph embeddings on non-Euclidean manifolds via adversarial learning.

The second contribution of this paper consists in two novel change detection tests (CDTs) operating on CCMs.
The first CDT monitors the geodesic distances of each embedded graph w.r.t. the sample Fr\'echet mean observed in the nominal regime of the process.
The resulting stream of distance values is processed by a CDT based on the Central Limit Theorem (CLT).
The second method considers embeddings lying on the CCMs, and builds on a novel CDT based on the CLT for Riemannian manifolds \cite{bhattacharya2018differential}.

We report a comparative analysis of the developed embedding and change detection mechanisms, by testing our architecture on both synthetic data and real-world applications. In particular, we consider: 1) the detection of ictal and pre-ictal phases in epileptic seizures, using functional connectivity networks extracted from intracranial electroencephalograpy data, and 2) the detection of hostility between two human subjects, using skeletal graphs extracted from video frames. We show that our methodology is able to effectively exploit the non-Euclidean geometry of CCMs to detect changes in graph streams, even in the case of extremely small changes, consistently outperforming baseline algorithms.
The use of an ensemble of CCMs and the parameter-free discriminator are shown to almost always outperform other configurations, on all problems taken into account.

\section{Related work}
\label{sec:related_work}

The problem of detecting changes in a graph-generating process is relatively unexplored in the literature, with most works focusing on networks with a fixed topology and without attributes \cite{holme2015modern}.
Literature reviews of existing approaches to detect changes, anomalies, and events in temporal networks can be found in \cite{akoglu2013anomaly,ranshous2015anomaly,akoglu2015graph}. 
Some notable contributions in this regard include the matrix-decomposition-based algorithm of \cite{sun2008less}, the change point methods of \cite{barnett2016change,peel2015detecting} for large-scale and correlation networks, and the block-model of \cite{wilson2016modeling} to monitor a co-voting network evolving over time.
More recently, Zambon \textit{et al.} \cite{zambon2017concept} proposed a theoretical framework allowing to face change detection problems on graph streams by considering embedding techniques. To the best of our knowledge, the work of Zambon \textit{et al.} \cite{zambon2017concept} is the first one addressing the problem by considering each graph in the stream as a random variable, hence allowing to perform change detection by means of classical, statistically motivated methods.

We note that none of the mentioned works applies modern deep learning methods to compute graph embeddings (i.e., represent a graph as a point in some geometric space), resorting to either feature extraction or classical dissimilarity-based embeddings.
To this end, and motivated by the contribution of our paper, in the following we introduce recent works on unsupervised learning of graph embeddings.

Focusing on recent literature regarding unsupervised deep learning on graphs and Riemannian manifolds \cite{bronstein2016geometric}, we mention that graph autoencoders (GAE) are typically used to encode the topological structure and node content of a single graph \cite{kipf2016variational, van2017graph, selvan2018extraction}; in this framework, an adversarially regularised GAE is proposed in \cite{pan2018adversarially}.
Closer to our approach, Simonovsky and Komodakis \cite{simonovsky2018graphvae} propose a graph variational autoencoder operating on batches of graphs rather than on a single graph. Their architecture focuses on variational inference for generating molecules, and adds a graph matching step between the input and reconstructed samples in order to support non-identified nodes. 
Several works in the literature introduce different approaches to model the latent space geometry of generative models, or study the geometry of the data distribution in order to facilitate the autoencoder in learning a non-Euclidean representation.
Davidson \textit{et al.} \cite{davidson2018hyperspherical} introduce a variational autoencoder based on the von Mises--Fisher distribution, aimed at modelling the spherical geometry underlying directional data. Korman \cite{korman2018autoencoding} proposes to use the AAE framework to recover the manifold underlying a data distribution, without making assumptions on the geometry of the manifold. This is achieved by approximating the manifold as a set of charts, each represented by the latent space of a linear AAE trained to match a uniform prior. The Riemannian geometry of deep generative models is also studied in \cite{shao2017riemannian,chen2017metrics}, whereas \cite{7439822} studies the metric-preserving properties of neural networks with random Gaussian weights.
In order to capture the hierarchical structure of domains like natural language, Nickel and Kiela \cite{nickel2017poincare} develop a technique based on stochastic gradient descent on manifolds for embedding graph data on a Poincar{\'e} ball.

\section{Background}
\label{sec:background}

\subsection{Stochastic processes in the space of graphs}
\label{sec:background-graphs}
A graph of order $N$ can be defined as a set of $N$ nodes and a set of pairwise relations, also called \textit{edges}, between them. Both nodes and edges can be associated with attributes.
We represent a graph as a 3-tuple of matrices $(A, X, E)$, where: 
\begin{itemize}
    \item $A$ is the binary adjacency matrix of size $N \X N$, where each element $A_{ij} \in \{0, 1\}$ encodes the presence or absence of an edge between nodes $i$ and $j$;
    \item $X$ is a matrix of size $N \X F$, where each row $X_i \in \R^F$ represents the $F$-dimensional real-valued attributes of node $i$;
    \item $E$ is a matrix of size $N \X N \X S$, where each element $E_{ij} \in \R^S$ represents the $S$-dimensional real-valued attributes of the edge between nodes $i$ and $j$ (if the edge is absent, we still include the entry as a vector of zeros).
\end{itemize}

We denote with $\mc G$ the set of all possible graphs as described above.
Attributes can encode all kinds of information, from numerical to categorical (e.g., via one-hot encoding).
In general, graphs in $\mc G$ can have different topology, order, and node or edge attributes. Moreover, we distinguish between \emph{identified} nodes, where there is a one-to-one correspondence between nodes of different graphs, and \emph{non-identified} nodes, where such a correspondence is absent or unknown.
In practice, the case of non-identified nodes is handled by using graph matching or graph alignment algorithms \cite{gm_survey,emmert2016fifty}.

In this paper, we consider a discrete-time stochastic process $\{g_t\}_{t \geq 0}$, where at each time step we observe a graph-valued random variable $g_t \in \mc G$. We refer to this process as a \emph{graph-generating process} and to the generated sequence as \emph{graph stream}.
In this setting, each observation $g_t$ can therefore differ from any other observation in terms of topology, order, and attributes. The considered setting differs from traditional graph signal processing, where the graph topology is usually assumed to be fixed and only the signal over the graph varies over time (see Section \ref{sec:related_work}). 

\subsection{Constant-curvature manifolds}
\label{sec:manifolds}

A CCM is a Riemannian manifold characterised by a sectional curvature $\kappa\in\R$ which is constant over the entire manifold. 
To each value of curvature $\kappa$, we associate a unique manifold $\M_\kappa$ whose geometry is fully determined by $\kappa$; 
in particular, three geometries emerge: spherical (positive curvature, $\kappa > 0$), flat (null curvature, $\kappa = 0$) and hyperbolic (negative curvature, $\kappa < 0$).

The special case of null curvature corresponds to the usual Euclidean space that is equipped with the ordinary Euclidean $\ell^2$-metric
\begin{equation*}
    \rho(x, y) = \norm{x - y}{2}.  
\end{equation*}

For $\kappa\ne 0$, the $d$-dimensional manifold $\M_\kappa$ can be represented using a $(d+1)$-dimensional real coordinate system, called the ambient space, and identified by the set of points
\begin{equation*}
    \left\lbrace x\in\R^{d+1}\;|\; \langle x, x \rangle_\kappa = \kappa^{-1}\right\rbrace,    
\end{equation*}

where $\langle \cdot, \cdot \rangle_\kappa$ is a scalar product that depends on the sign of $\kappa$:
for a positive curvature, $\langle x, y \rangle_\kappa = x^T y$ is the usual Euclidean inner product, whereas for a negative curvature
\begin{equation*}
    \langle x, y\rangle_\kappa = x^T \,
    \begin{pmatrix}
    I_{d\times d} & 0 \\ 0 & -1
    \end{pmatrix} \,y.
\end{equation*}

The associated geodesic metric, for $\kappa>0$, is
\begin{equation*}
    \rho(x, y) = \frac{1}{\sqrt{\kappa}}\arccos(\kappa\langle x, y \rangle_\kappa),    
\end{equation*}

and, for $\kappa<0$,
\begin{equation*}
    \rho(x, y) = \frac{1}{\sqrt{-\kappa}}\arccosh(\kappa\langle x, y \rangle_\kappa).    
\end{equation*}

We point out that there are other possible CCMs besides the ones considered here, e.g., cylinders.
However, here we consider the three aforementioned geometries (hyperspherical, flat, and hyperbolic) and leave the exploration of other CCMs as future research.

\subsection{Probability distributions on CCMs}
\label{sec:distribution_on_manifolds}

\begin{figure}
    \centering
    \includegraphics[width=0.7\linewidth, keepaspectratio=true]{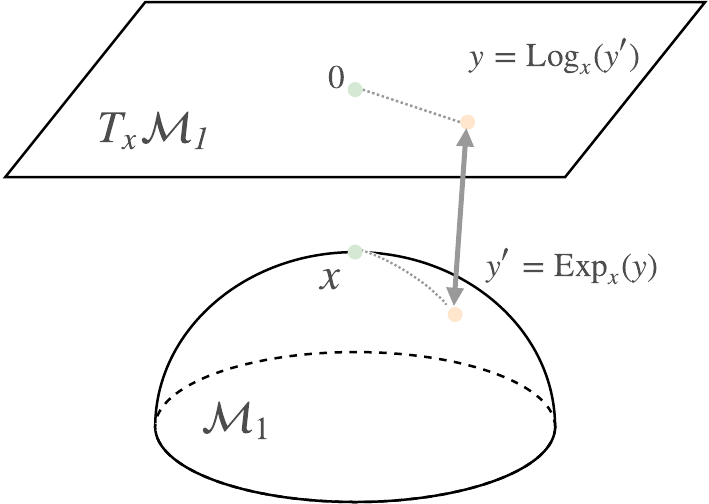}
    \caption{Schematic representation of the exp-map and log-map for $\M_1$, with a tangent space $T_x\M_1$ at $x$. We adopt a representation of the tangent plane such that the origin $x \in \M_\kappa$ of the log-map is mapped to the origin of the tangent plane (denoted $0$ in the figure) and vice versa with exp-map \cite{bridson2013metric}.}
    \label{fig:exp_map}
\end{figure}

Given a Riemannian manifold $\M$, and a tangent space $T_x \M$ at point $x$, we denote with $\Exp_{x}(\cdot)$ the Riemannian exponential map (exp-map), mapping points from the tangent space to the manifold, and with $\Log_{x}(\cdot)$ the logarithmic map (log-map), going from $\M$ to $T_x \M$ \cite{bridson2013metric,straub2015dirichlet}. 
The exp-map associates a point $y\in T_x\M$ with a point $y'\in\M$ so that the geodesic distance between $\Exp_x(y)$ and the tangent point $x$ equals the distance from $y$ to the origin of $T_x\M$ (Figure \ref{fig:exp_map}).
The log-map is defined as the inverse of the exp-map (on domain and co-domain of the exp-map) and has an analogous distance-preserving property. 
Note that, in general, the exp-map and log-map are defined only locally.
However, for every CCM $\M_\kappa$ of positive curvature, it is sufficient that $\langle x, y\rangle_\kappa \ne - \kappa$ in order for the log-map $\Log_x(y)$ to be well-defined. Conversely, when $\kappa\le 0$, the exp-map and log-map are defined in the entire tangent space and manifold, respectively.

Following \cite{straub2015dirichlet}, we use the exp-map operator to define a probability distribution with support on a CCM $\M_\kappa$ for $\kappa \ne 0$ (the case $\kappa=0$ is immediate, as both exp- and log-map correspond to the identity function).
In particular, given a probability distribution $P(\theta)$ on $T_x\M_\kappa$, parametrised by vector $\theta$, we consider the push-forward distribution $P_{\M_\kappa}(\theta)$ of $P(\theta)$ through $\Exp_x(\cdot)$, which can be interpreted as first sampling a point on $T_x\M_\kappa$ from $P(\theta)$, and then mapping it to $\M_\kappa$ using $\Exp_{x}(\cdot)$.
In this work, we always choose as origin of the exp-map the point $x \in \R^{d+1}$ with $x_i = 0, i=1, \dots, d$, and $x_{d+1} = |\kappa|^{-1/2}$.
Although this sampling procedure is suitable for CCMs with $\kappa \ne 0$, we keep the same subscript notation even for distributions with support on Euclidean spaces (denoted $P_{\M_0}(\theta)$).

\subsection{Adversarial autoencoders}
\label{sec:aae}

Adversarial autoencoders (AAEs) \cite{makhzani2016adversarial} are probabilistic models based on the framework of generative adversarial networks (GANs) \cite{goodfellow2014generative}. In AAEs, the encoder network of an autoencoder acts as the generator of a GAN, and is trained to match the aggregated posterior of its representation to an arbitrary prior distribution defined on the latent space. The training of an AAE is constituted of two phases: in the \textit{reconstruction} phase, both the encoder and the decoder networks are updated to minimise the reconstruction loss on the data space; in the \textit{regularisation} phase, a discriminator network is trained to distinguish between samples coming from the encoder and samples coming from the prior distribution, and finally the encoder network is updated to maximally confuse the discriminator.
The iterative repetition of these training steps results in a min-max game between the encoder and the discriminator \cite{goodfellow2014generative}, with both networks improving at their tasks (fooling and discriminating, respectively), until an equilibrium is reached and the aggregated posterior of the encoder is ideally indistinguishable from the true prior \cite{makhzani2016adversarial}. 
Some of the main advantages of AAEs are their modular architecture and flexibility in the choice of priors: they do not need an exact functional form of the prior in order to learn, unlike other probabilistic models like variational autoencoders.
In particular, we show in later sections how the discriminator network can be replaced with non-parametric functions to impose a geometrical regularisation on the latent space of the AAE, leading to a significantly reduced overall computational complexity.

\section{Change detection with graph embeddings on CCMs}
\label{sec:method}

We consider the task of determining whether the probability distribution underlying a graph-generating process has changed or not, from the nominal distribution $Q_0$ to a non-nominal one, $Q_1$.
Our methodology consists of training an AAE to compute graph embeddings on a CCM, and then exploiting the geometrical properties of the non-Euclidean embedding space to run a change detection test.

The algorithm is split between a training and an operational phase.
During the training phase, we observe a finite stream of graphs, $G_{train}$, coming from the nominal distribution $Q_0$.
The training stream is then mapped to the CCM using the encoder network, and a statistical analysis is performed there, in order to configure the CDT (details provided in Section \ref{sec:cdt}). In the operational phase, we monitor the graph-generating process, which is again mapped to the CCM using the encoder, with the aim of raising an alarm when a change in stationarity is detected. 
In the following sections, we give the details of both the embedding procedure on CCMs and the proposed CDTs.

\begin{figure}
    \centering
    \includegraphics[width=\linewidth, keepaspectratio=true]{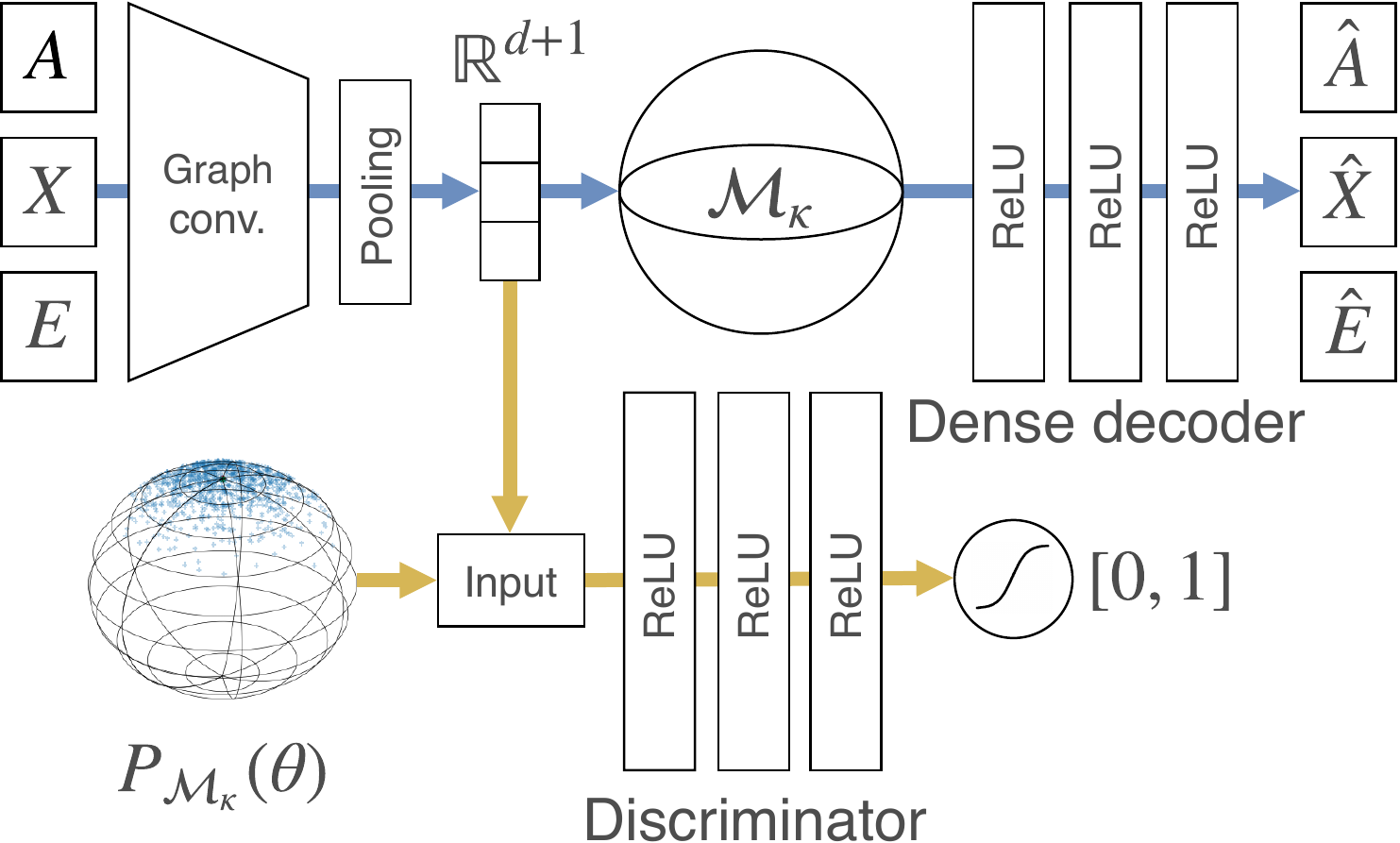}
    \caption{Schematic view of the AAE with a spherical CCM, in the probabilistic setting. From left to right, top to bottom: the AAE takes as input graphs represented by their adjacency matrix $A$, node features $X$, and edge attributes $E$, and outputs reconstructions of the same three matrices $\hat{A}$, $\hat{X}$, and $\hat{E}$ (blue path). The discriminator is trained to distinguish between samples produced by the encoder and samples coming from the true prior (yellow path). Finally, the encoder is updated to fool the discriminator by maximising its classification error. The encoder consists of graph convolutional layers, followed by a global pooling layer to obtain a graph embedding in the ambient space ($\R^{d+1}$). Embeddings are then constrained to the CCM $\mc M_k$ by 1) matching the prior $\mathcal{P}_{\M_\kappa}(\theta)$, and 2) orthogonally projecting the points onto the CCM. The decoder is a dense network with three parallel outputs for $\hat{A}$, $\hat{X}$, and $\hat{E}$. The discriminator is a dense network with sigmoid output. Best viewed in colour.}
    \label{fig:scheme}
\end{figure}

\subsection{Adversarial graph embeddings on CCMs}
\label{sec:adversarial_gae}

The proposed autoencoder (illustrated in Figure \ref{fig:scheme}) has a similar structure to the \textit{GraphVAE} network in \cite{simonovsky2018graphvae}, where each graph is mapped onto a point lying in the embedding space via global pooling.
Since all graph elements are taken into account for training both the encoder and the decoder of the AAE, the embeddings learned by the model represent all information available in the input graphs, from the topology to the attributes.

Graphs are represented with the matrix format described in Section \ref{sec:background-graphs}, using a 3-tuple $(A, X, E)$ for encoding both the topology and attributes.
The encoder network of the AAE is obtained by stacking graph convolutions \cite{defferrard2016convolutional, kipf2017semi, simonovsky2017dynamic}, which learn a representation of input graphs by transforming node features as a function of their neighbourhood. 
An advantage of using graph convolutions, w.r.t. more traditional methods like fully-connected networks, is that the computational complexity of such layers does not depend on the size of the input graphs, because they compute localised transformations of the nodes (not unlike the usual convolutional layers for images). This makes the encoder more efficient, and results in a low computational burden when deploying the model in real-world scenarios.
Moreover, because graph convolutions are invariant to node permutations, the encoder will produce the same embedding for any rotation of a given input graph. 

When edge attributes are present, we take them into account by using \textit{edge-conditioned} graph convolutions (ECCs) \cite{simonovsky2017dynamic} to learn the graph representation.
ECC layers $(l)$ compute a transformation $X^{(l)} \in \R^{N \X F_l}$ of the node attributes, from an input $X^{(l-1)} \in \R^{N \X F_{l-1}}$ as: 
\begin{equation}
\label{eq:ecc}
    X^{(l)}_{i} = \sum \limits_{j \in \mathcal{N}(i)} \frac{A_{ij}}{\sum\limits_k A_{ik}} \cdot X^{(l-1)}_{j} \cdot f(E_{ji}; \theta^{(l)}) + b^{(l)},
\end{equation}

where $f: \R^{S} \rightarrow \R^{F_{l-1} \X F_{l}}$ is a \textit{kernel-generating network}, parametrised by $\theta^{(l)}$, that outputs convolution kernels as a function of edge attributes, $b^{(l)}$ is a bias vector, and $\mc N(i)$ indicates the neighbours of node $i$.
Alternatively, when only node attributes are present, ECCs can be replaced with the graph convolution proposed by \cite{kipf2017semi}, which has an equivalent formulation but replaces the dynamic weighting of ECCs (i.e., the kernel-generating network) with a single trainable kernel $W \in \R^{F_{l} \X F_{l-1}}$, and adopts a normalised version of the graph Laplacian instead of the normalised adjacency matrix.
A graph-level pooling layer, like the global gated attention pooling one proposed in \cite{li2015gated}, is then used to aggregate the node-level embeddings in a single vector describing the graph globally. 
The latent space of the AAE is produced by $d+1$ neurons in the innermost layer (either by considering $d+1$ channels for gated pooling, or with a dedicated linear layer), and represents the ambient space of the target CCM.

Finally, the decoder network is a fully connected network that maps the latent representation to the graph space, by reconstructing $A$, $X$, and $E$. Such a formulation of the decoder introduces a noticeable computational cost, and puts a practical limit on the number of nodes that can be processed by the model, i.e., the maximum $N$ depends on the available memory in the GPU (e.g., for GraphVAE \cite{simonovsky2018graphvae} the authors reported results for $N \le 38$). Moreover, the dense decoder is not permutation-invariant and requires to perform graph alignment at training time when considering non-identified nodes.
For lack of a better solution in the literature, we resort to this formulation as in \cite{simonovsky2018graphvae}. However, we stress that the limit is only practical and that no theoretical limitation affects the method in this regard. Finally, this added complexity is only present at training time, because the decoder is not used to run the change detection test.
In the reconstruction phase, given an input graph $(A, X, E)$ and a reconstruction $(\hat{A}, \hat{X}, \hat{E})$, the model is trained to minimise
\begin{align}
\label{eq:loss}
    L = &- \frac{1}{N^{2}} \sum\limits_{i,j}\left( A_{ij} \log \hat{A}_{ij} + (1 - A_{ij}) \log (1 - \hat{A}_{ij})\right) + \\
      &+ \frac{1}{N} \sum\limits_{i} \norm{X_{i} - \hat{X}_{i}}{2}^2 + \frac{1}{N^{2}} \sum\limits_{i,j} \norm{E_{ij} - \hat{E}_{ij}}{2}^2 \nonumber
\end{align}

which consists of a cross-entropy loss for the binary adjacency matrix and mean squared error terms for the real-valued node and edge attributes. The loss function can easily be adapted to consider categorical or binary attributes by choosing an appropriate loss function for the corresponding matrix. Note that the three terms can be multiplied by a scalar weight to control their importance in the resulting loss \eqref{eq:loss}; here, we follow \cite{simonovsky2018graphvae} and weight each term equally.

We train the AAE on a sequence $G_{train}$ of nominal graphs, conditioning its aggregated posterior to match the true prior $P_{\M_\kappa}(\theta)$. 
The prior implicitly defines the geometric constraint that we wish to impose on the latent space, so that the representation on the CCM can be autonomously learned by the AAE in order to fool the discriminator. 

The discriminator $D(\vec z)$ is a neural network computing the likelihood that a point $\vec z \in \R^{d+1}$ is a sample from the prior $P_{\M_\kappa}(\theta)$, rather than an embedding produced by the encoder \cite{goodfellow2014generative}. We train the discriminator using samples from $P_{\M_\kappa}(\theta)$ as positive examples, and embeddings from the encoder as negative examples. The encoder is then updated by backpropagating the loss's gradient through the discriminator, using graphs from the data distribution as positive examples.

Since the training procedure imposes only a soft constraint on the latent representation, there are no guarantees that all embeddings will exactly lie on the CCM, making it impossible to compute exact geodesic distances between embeddings.
To compensate for this issue, when running the change detection tests we orthogonally project the embeddings onto the CCM. The projection is also included during the training reconstruction phase, immediately before the decoder (see Figure \ref{fig:scheme}). This does not impact the regularisation of the encoder network, but pushes the decoder to learn a meaningful map from the CCM to the graph space.
Note that it would be possible to directly project the embeddings onto the CCM without adversarial regularisation. However, empirical results (not shown) indicate how this would significantly compromise the performance in terms of representation and, most importantly, change detection. 

\subsection{Geometric discriminator}
\label{sec:geom_discrim}

Enforcing a distribution on the latent representation is not our primary goal, as the key element of the proposed architecture is the geometric regularisation of the encoder's latent space. Moreover, imposing a prior on the latent representation could in principle interfere with the statistical analysis preformed by the CDT, and introduce unwanted effects in the behaviour of the algorithm.
Therefore, we propose a variation of the AAE that replaces the implicit regularisation based on matching the prior distribution, with an explicit regularisation term imposed by a parameter-free discriminator, used to compute the membership degree of an embedding to the CCM.
By maximally fooling this \textit{geometric} discriminator, the encoder is explicitly optimising its representation to lie on the target CCM. Moreover, replacing the discriminator network with a parameter-free model gives an advantage on those problems characterised by a scarcity of data, like the seizure detection task detailed in Section \ref{sec:seizure_detection}.
Since the geometric discriminator does not need to be trained, we skip the first step of the regularisation phase and only update the encoder in the final step of the training loop.

For a CCM $\M_{\kappa}$ with $\kappa \ne 0$, the non-parametric discriminator is defined as:
\begin{equation}
    \label{eq:geom_discriminator}
    D_{\M_{\kappa}}(\vec z) =  \mathrm{exp}\left(\cfrac{-\big( \langle \vec z, \vec z \rangle_\kappa - \frac{1}{\kappa} \big)^2}{2\varsigma^2}\right)
\end{equation}

where $\varsigma$ is a hyperparameter that controls the width of the membership function.
Equation \ref{eq:geom_discriminator} defines the membership degree of $\vec z$ to the CCM, where $D_{\M_\kappa}(\vec z) = 1$ when the embedding lies exactly on $\M_\kappa$, and $D_{\M_\kappa}(\vec z) \rightarrow 0$ when it is far away.
When $\kappa = 0$, the CCM corresponds to the entire latent space (c.f.\ Section \ref{sec:manifolds}) and the geometric discriminator outputs 1 for all points. In this case, the formulation of the network is equivalent to the standard autoencoder, because during the regularisation phase the encoder is not updated (the loss is always 0). 
\subsection{Change detection on CCMs}
\label{sec:cdt}

The general test hypotheses considered for detecting a change in stationarity in the distribution of an i.i.d. graph stream $g_1,g_2,\dots,g_i,\dots$, observed during the operational phase, are
\begin{equation}
    \begin{aligned}
    H_0&: g_i\sim Q_0,\ i=1,2,\dots\\
    H_1&: g_i\sim
    \begin{cases}
    Q_0 & \ i<\tau\\
    Q_1 & \ i\geq\tau,
    \end{cases}
    \end{aligned}
\end{equation}

where $\tau$ indicates the point in the sequence where the change occurs. 
$Q_0$, $Q_1$, and $\tau$ are unknown.
During the operational phase, we use the encoder network to convert the incoming graph stream into a multivariate stream of embeddings $\vec z_i \in \M_\kappa$, which is then monitored by a sequential statistical test to detect a possible change in the nominal distribution. Accordingly, the graph stream $G_{train}$, on which we trained the AAE, is converted to a stream of embeddings $Z_{train}$. 

Our change detection methodology builds on the CDT proposed by \cite{zambon2017concept}, by extending it to the case of CCMs. 
More in detail, the CDT considers a generic stream of vector points $u_1,u_2,\dots,u_i,\dots$, which is processed in windows of $n$ points at a time, so that for each $w=1,2,3,\dots$, a window $[u]_w$ containing $u_{(w-1)n+1},\dots,u_{wn}$ is generated, and a statistic $S_w$ is computed by means of the accumulation process typical of the cumulative sums (CUSUM) chart \cite{page1954continuous}. 
Statistic $S_w$ has a global role, as it recurrently accumulates information from \textit{local} statistics $s_i=s([u]_i)$, for $i = 1, \dots, w$, as 
\begin{equation*}
    S_w = \max\{0, S_{w-1} + s_w - q\},
\end{equation*}

with $S_0=0$ and $q$ a parameter tuning the sensitivity of the test.
The null hypothesis $H_0$ is rejected any time $S_w$ exceeds a threshold $h_w$, and the algorithm raises an alarm indicating that a change has been detected; the accumulator $S_w$ is then reset to 0.
After the first alarm is raised, the change point is estimated as 
\begin{equation*}
    \hat \tau=n\cdot\min\{ w | S_w > h_w \}.
\end{equation*}

Threshold $h_w$ is set according to a user-defined significance level $\alpha$, by requiring, under the null hypothesis $H_0$, that %
\begin{equation*}
    \mathbb{P}(S_w > h_w | H_0, S_i  \leq h_i, i < w) = \alpha.
\end{equation*}

The threshold is set so that the probability of having a false alarm at generic step $w$ is $\alpha$, hence allowing us to control the false positive detection rate.

Note that the scoring function $s_w = s([u]_w)$ entirely defines the behaviour of the CDT, and that by knowing the distribution of $s_w$ we can compute the threshold $h_w$ given $\alpha$. Here, we consider $s_w$ to be the Mahalanobis distance
\begin{equation}
    \label{eq:mahalanobis}
    s_w = (\mathbb{E}[u] - \overline{[u]}_w)^T {\rm Cov[u]}^{-1} (\mathbb{E}[u] - \overline{[u]}_w),
\end{equation}

between the sample mean $\overline{[u]}_w$ of $[u]_w$ and the expected value $\mathbb{E}[u]$ of $u$. In the stationary case, thanks to the CLT, it can be shown that $n \cdot s_w \sim \chi^2$.

We propose two different ways of computing the points $u_i$, both exploiting the geometry of the CCMs. By monitoring the mean of the sequence, we are able to detect changes in the distribution driving the graph-generating process. Since we use graph convolutions in the encoder network, changes in the distribution of $A$, $X$, and $E$ are all reflected on the embeddings (c.f.\ Equation \ref{eq:ecc}), and can therefore be detected by the CDTs.

\subsubsection{Distance-based CDT (D-CDT)} the first proposed CDT considers the nominal distribution $F_0$ of the training stream of embeddings, derived as the push-forward distribution of $Q_0$ through the encoder network. 
The Fr\'echet mean of $F_0$, denoted as $\mu_0$, is estimated over the training sequence $Z_{train}$ as
\begin{equation}
    \label{eq:frechet-mean}
        \mu_0 =\argmin_{\vec z \in \M_\kappa} \sum\limits_{\vec z_i \in Z_{train}} \rho(\vec z_i, \vec z)^2,
\end{equation}

where $\rho(\cdot, \cdot)$ is the geodesic distance as defined in Section \ref{sec:manifolds}.

For each embedding $\vec z_i \in \M_\kappa$ in the operational stream, then, we consider $u_i=\rho(\mu_0,\vec z_i)$.  The resulting sequence $u_1,u_2,\dots, u_i,\dots$ is finally monitored with the CDT presented above.

\subsubsection{Riemannian CLT-based CDT (R-CDT)} our second implementation of the CDT builds on a Riemannian version of the CLT proposed in \cite{bhattacharya2018differential}, which adapts the Mahalanobis distance \eqref{eq:mahalanobis} to non-Euclidean manifolds. In this case, the operational stream of embeddings $\vec z_i \in \M_\kappa$ is mapped to the tangent space $T_{\mu_0}\M_\kappa$ with
\begin{equation*}
    u_i = \Log_{\mu_0}(\vec z_i),    
\end{equation*}

and the usual CDT is applied using the modified local statistic $s_w$.
In the case of $\kappa = 0$, the standard CLT applies directly to the embeddings without modifying $s_w$.

\subsection{Setting CDT parameters} 
\label{sec:cdt_params}

In the literature, e.g. \cite{montgomery2007introduction}, it is suggested as a good practice to set $q$ as half of the increase in $\mathbb{E}[s_w]$ that the designer expects to observe.
It is possible to show that the change detection procedure can identify any change of magnitude larger than $q$, independently from significance level $\alpha$: to every $\alpha$ in fact is associated a threshold $h$, and the expected time of detection is 
\begin{equation*}
    \hat t < \frac{h}{(\mathbb{E}[s_w|H_1] - q)},
\end{equation*}

where $\mathbb{E}[s_w|H_1]$ is the expected value of $s_w$ in the non-nominal conditions.
Although in principle it is possible to detect arbitrarily small shifts by setting $q = \mathbb{E}[s_w|H_0]$, we suggest to avoid this setting because any (even small) bias introduced at training time in estimating $\mathbb{E}[s_w|H_0]$ will eventually trigger a false alarm.

Parameter $\alpha$ corresponds to type-I errors of the statistical test, that is, to the probability of rejecting $H_0$ when $H_0$ is known to be true. Parameter $\alpha$, is therefore directly related to the rate of false alarms, with smaller values of $\alpha$ corresponding to fewer false alarms. However, we note that a smaller $\alpha$ corresponds also to larger delays of detection under $H_1$. 
Depending on the application at hand, the user should determine the best trade-off and the rate of false alarms that can be tolerated.

Finally, the size $n$ of the windows processed by the CDT should be large enough to consider $n\cdot s_w\sim \chi^2$, thus yielding the desired significance level $\alpha$. However, processing larger windows of observations in the operational phase of the algorithm will result in a lower time resolution. 

\subsection{Ensemble of CCMs}
\label{sec:ensemble_ccms}

In most applications, we do not have prior information about the optimal CCM for embedding the data distribution, and choosing the optimal CCM for a specific task may not be trivial. Therefore, here we propose to use an ensemble of CCMs, each characterised by a different curvature.
The ensemble of CCMs is denoted using the product space notation as $\M_* = \M_{\kappa_1} \X \dots \X \M_{\kappa_i} \X \dots \X \M_{\kappa_c}$. In practice, we consider each manifold separately, and the AAE is trained to optimise the latent representation in parallel on each CCM.
Adapting the AAE to the ensemble case is as simple as considering $c$ parallel fully connected layers after pooling, each producing a representation in a $(d+1)$-dimensional ambient space; when $\kappa=0$, we assume that $\M_0$ has dimension $d+1$, rather than $d$.
The produced embeddings are then concatenated in a single $c(d+1)$-dimensional vector before being fed to the discriminator.
Similarly, the prior is defined as the concatenation of $c$ samples $\vec z_i \sim P_{\M_{\kappa_i}}(\theta)$, one for each CCM. 
When using the geometric discriminator \eqref{eq:geom_discriminator}, given an embedding $\vec z = [\vec z_1| \dots| \vec z_c]^T$, we apply the geometric classifier $D_{\M_{\kappa_i}}$ on each CCM, and compute the average membership as 
\begin{equation*}
    D_{\M_*}(\vec z) = \frac{1}{c}\sum_{i = 1}^{c} D_{\M_{\kappa_i}}(\vec z_i).    
\end{equation*}

The orthogonal projection of the embeddings is also performed separately on each CCM.

Accordingly, we also adapt the CDTs described in Section ~\ref{sec:cdt} to consider the ensemble of CCMs.
For D-CDT, we compute for each CCM the same distance-based representation as in the single CCM case.
This results in a multivariate stream of $c$-dimensional vector of distances, which can be monitored by the base CDT. 
Similarly, the CDT operating on manifolds is adapted by considering a R-CDT for each CCM $\M_{\kappa_i}$. The ensemble of statistical tests raises an alarm any time at least one of the individual tests detects a change. Since the tests are in general not independent, we apply a Bonferroni correction \cite{bonferroni1936teoria} to each R-CDT, so that the overall significance level is at least the user-defined level $\alpha$.

\section{Experiments}
\label{sec:experiments}

To test our methodology, we consider three different application scenarios.
First, we evaluate the performance of our model on a synthetic stream of Delaunay triangulations where we are able to control the difficulty of the change detection problem.
Second, we consider two intracranial electroencephalograpy (iEEG) datasets for epileptic seizure detection and prediction, characterised by changes of different magnitudes.
Finally, we consider a computer vision problem, using the NTU RGB+D skeleton dataset for action recognition \cite{Shahroudy_2016_CVPR}, with the aim of detecting changes from a non-violent to a violent interaction between two subjects.

\subsection{Experimental setting}
\label{sec:exp-setting}

We test our methodology by considering three different CCMs, namely the Euclidean $\M_0$, hyperspherical $\M_1$, and hyperbolic $\M_{-1}$ manifolds. For $\M_1$ and $\M_{-1}$, we take $d=2$ and, accordingly, a three-dimensional ambient space.
By choosing a low-dimensional manifold, we encourage the encoder to learn an abstract representation of the graphs, and in particular we are also able to visualise the representation learned by the network for a qualitative assessment of the algorithm (e.g., Figures \ref{fig:embeddings_delaunay} and \ref{fig:embeddings_action}).
For $\M_0$, we keep the structure of the autoencoder unchanged and consider a three-dimensional latent space.
Since we are unable to identify \textit{a priori} the best curvature for the problems taken into account, we also consider an ensemble composed of all three geometries, $\M_* = \M_{-1} \X \M_{0} \X \M_{1}$.
Note that the specific values of $\kappa$ are only important for their sign, which determines the geometry of the CCMs. Since we are not interested in imposing any other constraint on the representation (e.g., minimising the distortion introduced by the embedding process \cite{zambon2018anomaly}), the magnitude of the curvature can safely be ignored, as it only affects the scale of the representation. Thus, we choose $\kappa = -1, 0, 1$ to simplify the implementation of the experiments. 
We learn a representation on each manifold using both the probabilistic AAE formulation and the non-parametric geometric discriminator \eqref{eq:geom_discriminator}. For each type of embedding, we run both D-CDT and R-CDT.

The architecture of the AAE is closely inspired to \textit{GraphVAE} \cite{simonovsky2018graphvae}, and we conduct a brief hyperparameter search for each experiment, using the validation loss of the network for model selection. 
Like in \cite{simonovsky2018graphvae}, the encoder consists of two graph convolutional layers with 32 and 64 channels respectively, with batch normalisation, ReLU, and L2 regularisation (with a factor of $5\cdot 10^{-4}$), followed by global attention pooling with 128 channels. When using ECC layers, the kernel-generating network consists of two fully connected ReLU layers of 128 units, with a linear output of $F_l \cdot F_{l-1}$ neurons. The latent representation is produced by a ReLU layer with 128 units followed by a linear layer with $d+1$ units (these last two layers are replicated in parallel when considering the ensemble of CCMs). The decoder is a fully connected three-layer network of 128, 256, and 512 neurons, with ReLU and batch normalisation, followed by three parallel output layers to reconstruct the graphs: a sigmoid layer for $A$, and two layers for $X$ and $E$, with activations according to their specific domain (e.g., for categorical attributes we could use a softmax activation).

We consider a discriminator network with three hidden ReLU layers of 128 units each.
For the prior, we consider the commonly used Gaussian distribution $\mathcal{N}_{\M_{\kappa_i}}(0, 1)$, adapted to have support on the CCM (c.f.\ Section \ref{sec:distribution_on_manifolds}). When using the geometric discriminator we set $\varsigma = 5$.
We train all networks using Adam \cite{kingma2014adam} with a learning rate of $0.001$ and a batch size of 128. We train to convergence, monitoring the validation loss with a patience of 20 epochs. We set aside 10\% of the samples for testing, and 10\% for validation and model selection.
For each graph, $X$ and $E$ are normalised element-wise, by removing the mean and scaling to unit variance.
For the CDTs we set $\alpha = 0.01$ and $q$ to the $0.75$ quantile of the $\chi^2$ distribution of $s_w$. The size of the windows processed by CUSUM is set to $0.1\%$ of the number of training samples, which we found to be enough to estimate the mean and variance of $s_w$ in Equation \ref{eq:mahalanobis}. 

The reference baseline is that of \cite{zambon2017concept}, for which we use the open-source implementation published by the authors\footnote{https://github.com/dan-zam/cdg}; there, we use a $(d+1)$-dimensional dissimilarity representation for the embedding.

\subsection{Performance metric for CDTs}
\label{sec:cdt_metric}

To evaluate the detection performance of a CDT, we consider the predictions of the algorithm (i.e., whether or not it is raising an alarm) for each point of the operational stream, and compare them with the ground truth (i.e., whether or not a change has actually occurred at a given time). 
In this setting, accuracy is not a fair performance indicator for the proposed CUSUM-based algorithms, because the detection delay of the CDT (due to the accumulation process) may result in low true positive rates even if the change is consistently detected by the algorithm.
To avoid this issue, we consider the \textit{run lengths} (RLs) of the CDT, defined as the number of time-steps between any two consecutive alarms. In the nominal regime, the CDT is configured to have a false positive rate of $\alpha$, and accordingly the average RL is $\sim 1/\alpha$. Conversely, in the non-nominal regime the detection rate should be significantly higher (ideally 1), and the average RL should be lower than the one under the nominal distribution. 
Therefore, by comparing the distributions of RLs in the two regimes, we are able to quantify the performance of the CDT. 

We test whether nominal RLs are statistically larger than non-nominal ones according to the Mann-Whitney U test \cite{mann1947test}. 
The resulting $U$ statistic is then normalised to obtain the Area Under the receiver operating characteristic Curve (AUC) score, which in our case measures the separability of the two RL distributions,
\begin{equation}
    AUC_{RL} = \cfrac{U}{N_0 N_1},
\end{equation}
where $N_0$ and $N_1$ are the sample sizes of the observed RLs in the two regimes, respectively.
This metric allows us to compare different algorithms operating on the graph streams, and is easy to compute starting from the alarms raised by the CDTs over time.

\subsection{Delaunay triangulations}
\label{sec:delaunay}

\begin{figure}
    \centering
    \includegraphics[width=\linewidth, keepaspectratio=true]{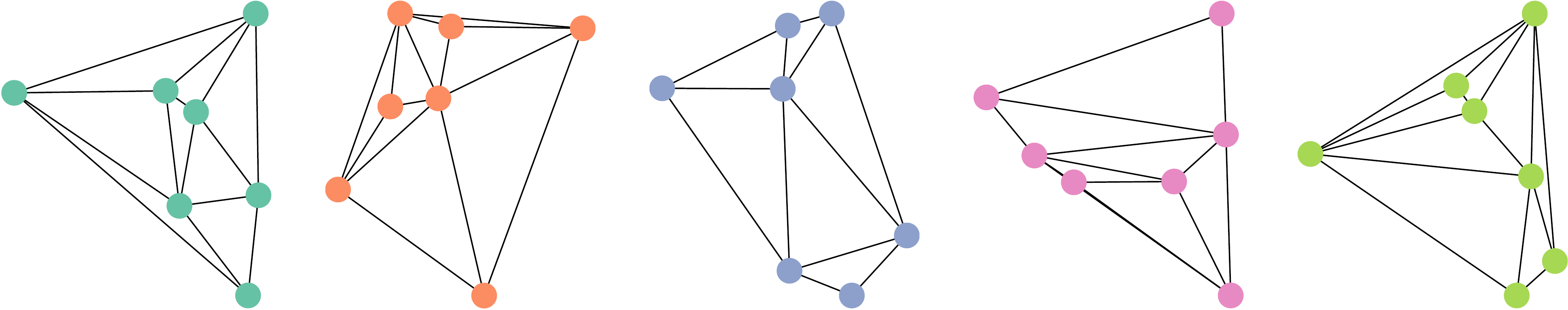}
    \caption{Example of Delaunay triangulations of classes 0 to 4. The same colours are used in Figure \ref{fig:embeddings_delaunay} to represent the embeddings on the CCMs.}
    \label{fig:delaunay_graphs}
\end{figure}

As a first experimental setting, we consider a synthetic graph stream of Delaunay triangulations, computed from a set of noisy observations on a 2D plane. 
Each observation is associated to a node, and the 2D coordinates are used as node attributes (i.e., $F=2$). The topology of the graphs is obtained from the Delaunay triangulation of the points, by adding an edge between two nodes if they are adjacent in a triangle (e.g., see Figure \ref{fig:delaunay_graphs}).

In particular, a class of graphs is defined starting from a fixed set of $N$ support points, and graph instances are generated by adding Gaussian noise to the support. By changing the support, we are able to generate different classes of graphs.
We wish to distinguish between the nominal class 0, and the non-nominal classes $C \geq 1$.
To generate a graph stream, we sample graphs of class 0 representing the nominal regime, and simulate a change by transitioning to a different class $C$ for the non-nominal regime. This allows us to have a ground truth with a known change point.

\subsubsection*{Data generation}
the support points of the nominal class 0 are sampled from a uniform distribution in $[0, 10]^2$. 
The support of non-nominal classes is then generated from the support of class 0, by adding to each point a vector sampled on the circumference of radius
\begin{equation}
    r^{(C)} = 10\biggr(\frac{2}{3}\biggr)^{C - 1}, 
\end{equation}
such that the support points of class $C$ are: 
\begin{equation}
    X^{(support, C)} = X^{(support, 0)} + r^{(C)} [\cos(\theta), \sin(\phi)], 
\end{equation}
where $\theta, \phi \sim \mc U (0, 2\pi)$ are random vectors in $\R^N$. 
Note that $\cos(\theta)$ and $\sin(\phi)$ are evaluated element-wise, so that $r^{(C)} [\cos(\theta), \sin(\phi)]$ is an $N \X 2$ matrix where each row is a random point on the circle. 

Graph instances of each class $i$ are generated by perturbing the support points with Gaussian noise:
\begin{equation}
    \label{eq:delaunay-class0}
    X = X^{(support, i)} + X^{(noise)},
\end{equation}

where $X^{(noise)}$ is a random matrix in $\R^{N \X F}$ of normally distributed components $\sim \mc N(0,1)$.

Intuitively, class indices are proportional to the difficulty of detecting a change, because the perturbations to the support get smaller as $C$ increases, making it more difficult to distinguish class $C$ from class $0$.

For the experiments, we generate $5\cdot10^3$ graphs of class 0 for the training stream $G_{train}$, and consider different operational streams with $C=1, \dots, 20$ for evaluating the performance on increasingly difficult problems. Each operational stream consists of $2\cdot10^4$ graphs, with change point $\tau = 10^4$, from class 0 to class $C$. We consider graphs of fixed order $N=7$.

\subsubsection*{Results} 

\begin{figure*}
    \centering
    \includegraphics[width=\textwidth, keepaspectratio=true]{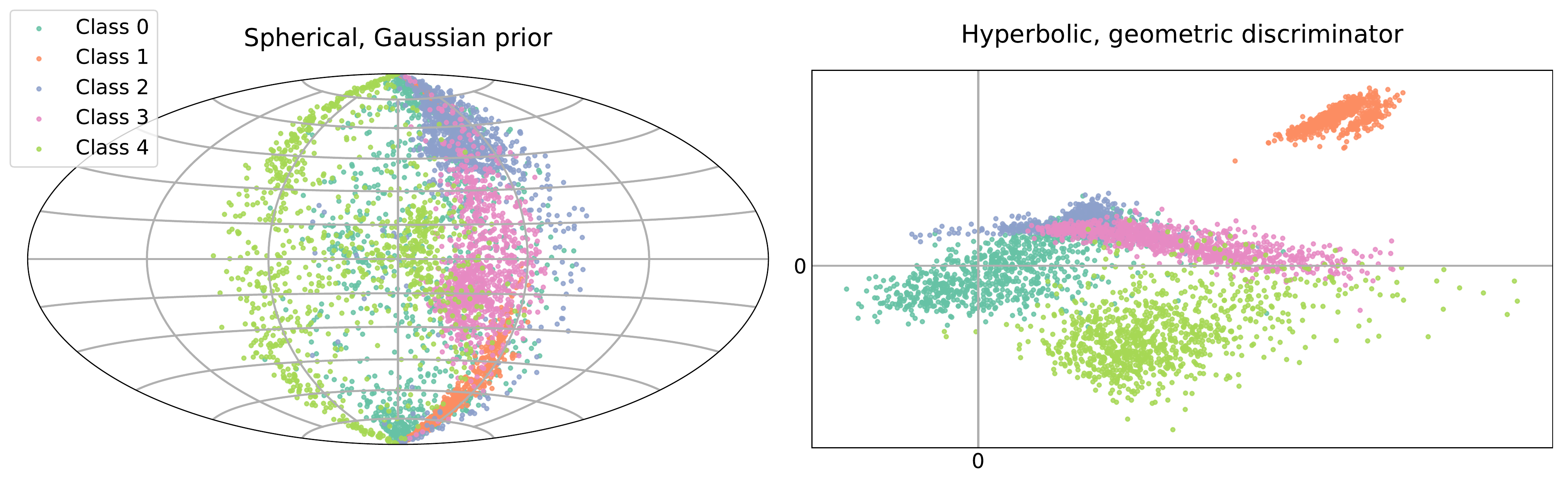}
    \caption{Hammer and planar projections of embeddings produced by AAEs with latent CCMs $\M_1$ and $\M_{-1}$, respectively, for Delaunay triangulations of classes 0 to 4 (c.f.\ Figure \ref{fig:delaunay_graphs}). Best viewed in colour.}
    \label{fig:embeddings_delaunay}
\end{figure*}

we report in Table \ref{tab:delaunay} the results obtained with each configuration of CCM, embedding type, and CDT. Results show that the combination of geometric discriminator and R-CDT on $\M_*$ consistently outperforms all other methods. This suggests that CCMs with different curvatures encode different yet useful information, which the algorithm is able to exploit. The representations learned by the AAE on different CCMs are shown in Figure \ref{fig:embeddings_delaunay}.
Moreover, we note a considerably high performance in those problems characterised by a less evident change, where the algorithm is able to detect node perturbations in the order of $10^{-3}$ (class 20).
We also note that, while performing comparably to the probabilistic AAE formulation in different configurations, the geometric discriminator still provides noticeable benefits on model complexity, particularly for smaller $N$ (as the complexity of the AAE is quadratic w.r.t. $N$). Here, for instance, we notice a significant reduction of up to $13.35\%$ in the total number of parameters (i.e., from $\approx252$k to $\approx218$k) w.r.t. using the standard discriminator.

\begin{table*}
\centering
\caption{AUC score on Delaunay triangulations. We report the results for even-numbered classes $C$ in the non-nominal regime. The \textit{CCM} column indicates the manifold on which the AAE is trained, whereas the \textit{CDT} column indicates the type of CDT used. In the \textit{Emb.} column, \textit{Prior} indicates that the embeddings were computed using the standard AAE probabilistic setting, while \textit{Geom} indicates the AAE with geometric discriminator \eqref{eq:geom_discriminator}. The baseline by \cite{zambon2017concept} is denoted as using $\M_0$ and R-CDT because it is formally equivalent to a R-CDT on Euclidean manifolds. We report the best results in bold.}
\label{tab:delaunay}
\begin{tabular}{lll|cccccccccc} 
\toprule
& & & \multicolumn{10}{c}{$C$} \\
\textbf{CCM} & \textbf{CDT} & \textbf{Emb.} & \textbf{20} & \textbf{18} & \textbf{16} & \textbf{14} & \textbf{12} & \textbf{10} & \textbf{8} & \textbf{6} & \textbf{4} & \textbf{2} \\  
\midrule
\multirow{4}{*}{$\M_*$}& \multirow{2}{*}{D-CDT} & Geom. & 0.49 & 0.50 & 0.51 & 0.64 & 0.56 & 0.51 & 0.91 & 0.79 & \textbf{1.00} & \textbf{1.00} \\
  & & Prior & 0.52 & 0.52 & 0.52 & 0.58 & 0.53  & 0.50 & 0.94 & 0.95 & \textbf{1.00} & \textbf{1.00}  \\
  & \multirow{2}{*}{R-CDT} & Geom. & \textbf{0.70} & \textbf{0.71} & \textbf{0.74} & \textbf{0.76} & \textbf{0.78} & \textbf{0.73} & \textbf{0.98} & \textbf{0.96} & \textbf{1.00} & 1.00 \\
  & & Prior & 0.42 & 0.24 & 0.32 & 0.21 & 0.28 & 0.20 & 0.55 & 0.72 & \textbf{1.00} & \textbf{1.00} \\ 
\midrule
\multirow{4}{*}{$\M_{-1}$} & \multirow{2}{*}{D-CDT} & Geom. & 0.49 & 0.55 & 0.54 & 0.61 & 0.59 & 0.48 & 0.71 & 0.53 & 0.96 & 0.90  \\
   & & Prior & 0.52 & 0.55 & 0.49 & 0.63 & 0.55 & 0.43 & 0.61 & 0.75 & \textbf{1.00} & 0.96  \\
   & \multirow{2}{*}{R-CDT} & Geom. & 0.47 & 0.52 & 0.55 & 0.64 & 0.59 & 0.60 & 0.91 & 0.95 & 0.99 & \textbf{1.00}  \\
   & & Prior & 0.48 & 0.53 & 0.54 & 0.63 & 0.58 & 0.39 & 0.96 & 0.55 & \textbf{1.00} & 0.93  \\  
\midrule
\multirow{4}{*}{$\M_0$} & \multirow{2}{*}{D-CDT} & Geom. & 0.55 & 0.45 & 0.50 & 0.73 & 0.60  & 0.47 & 0.31 & 0.40 & 0.99 & 0.99  \\
   & & Prior & 0.54 & 0.61 & 0.51 & 0.66 & 0.57 & 0.58 & 0.63 & 0.67 & \textbf{1.00} & 0.93  \\
   & \multirow{2}{*}{R-CDT} & Geom. & 0.53 & 0.54 & 0.54 & 0.56 & 0.60 & 0.58 & 0.65 & 0.62 & 0.94 & 0.94  \\
   & & Prior & 0.49 & 0.51 & 0.49 & 0.52 & 0.50 & 0.43 & 0.71 & 0.66 & 0.96 & 0.96  \\  
\midrule
\multirow{4}{*}{$\M_{1}$}   & \multirow{2}{*}{D-CDT} & Geom. & 0.37 & 0.53 & 0.45 & 0.48 & 0.48 & 0.53 & 0.68 & 0.83 & 0.76 & 0.90  \\
   & & Prior & 0.47 & 0.55 & 0.49 & 0.61 & 0.53 & 0.51 & 0.59 & 0.72 & 0.90 & 0.89  \\
   & \multirow{2}{*}{R-CDT} & Geom. & 0.42 & 0.53 & 0.49 & 0.53 & 0.55 & 0.56 & 0.85 & 0.80 & 0.92 & \textbf{1.00}  \\
   & & Prior & 0.45 & 0.50 & 0.53 & 0.59 & 0.55 & 0.50 & 0.79 & 0.77 & \textbf{1.00} & 0.99  \\  
\midrule
$\M_0$ & R-CDT & \cite{zambon2017concept} & 0.49 & 0.50 & 0.50 & 0.50 & 0.50 & 0.52 & 0.56 & 0.88 & 0.93 & 0.93  \\ 
\bottomrule
\end{tabular}
\end{table*}

\subsection{Seizure detection}
\label{sec:seizure_detection}

\begin{figure}
    \centering
    \includegraphics[width=\linewidth, keepaspectratio=true]{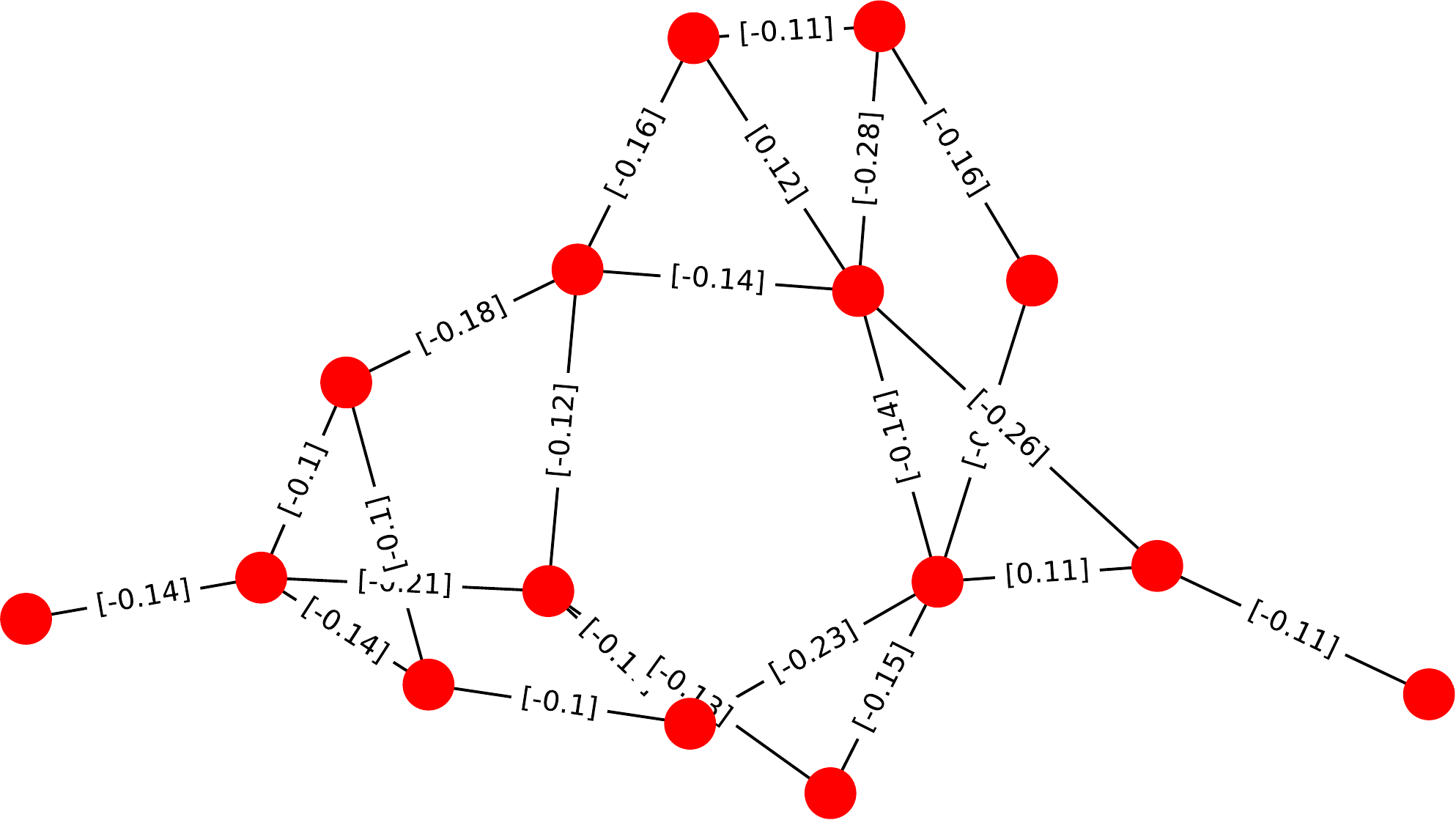}
    \caption{Example of functional connectivity network extracted from a 1-second clip of iEEG data, using Pearson's correlation. Only edge attributes are shown in the figure.}
    \label{fig:ieeg_graph}
\end{figure}

As a first real-world application to test our methodology, we consider iEEG data from Kaggle's \textit{UPenn and Mayo Clinic's Seizure Detection Challenge}\footnote{https://www.kaggle.com/c/seizure-detection} (SDC) and the \textit{American Epilepsy Society Seizure Prediction Challenge}\footnote{https://www.kaggle.com/c/seizure-prediction} (SPC).
A summary of the SDC and SPC datasets is provided in Table \ref{tab:ieeg-data}.
In these datasets, iEEG signals are provided as one-second clips belonging to two different classes, namely the nominal \textit{interictal} samples and the non-nominal \textit{ictal} (or \textit{preictal}, in the SPC case) samples. The datasets are collected from dogs and human patients, with a variable number of sensors applied to each patient, resulting in multivariate streams of different dimensions. Here, for the SDC datasets we consider only subjects with more than 1000 labelled clips, while for SPC we consider those with more than 500 (due to the datasets of SPC being overall smaller, with some patients having as little as 42 labelled clips).
Functional connectivity networks are widely used in neuroscience \cite{bastos2016tutorial} to represent the coupling between activity recorded from macro-units in the brain. Functional networks find a natural representation as weighted graphs, giving rise to a stream of attributed graphs with varying topology and attributes (see Figure \ref{fig:ieeg_graph} for an example), hence making the problem a suitable case study to test the proposed CDT methodology.

The training and operational graph streams are generated for each patient, using the labelled training clips in the datasets. We generate arbitrarily long streams via bootstrapping, sampling interictal graphs for the nominal regime, and ictal (or preictal) graphs in the non-nominal regime.

Graphs are generated from each one-second multivariate stream. As initial preprocessing step, we remove the baseline 60~Hz introduced by the recording devices with a Butterworth filter. The number of nodes $N$ in the graphs corresponds to the number of channels $N$ in the stream (see Table \ref{tab:ieeg-data}). We generate edge attributes using a functional connectivity measure estimated in the high-gamma band (70-100 Hz), such that $E \in \R^{N \X N}$ (i.e., $S=1$). We report experimental results using two different measures: 1) Pearson correlation and 2) the Directed Phase Lag Index (DPLI) \cite{bastos2016tutorial}. 
Finally, in order to encode information about each individual channel in the node attributes, we consider the first four wavelet coefficients ($F=4$) computed by means of discrete wavelet transform of the related signals \cite{bastos2016tutorial}.
As a final preprocessing step, we remove all edges with absolute value $ \le 0.1$, in order to have a non-trivial topology in the graphs (which otherwise would simply be fully connected, reducing the effectiveness of the graph convolutions).
Once again, we consider a training stream of $5\cdot10^3$ graphs, and an operational stream of $2\cdot10^4$ graphs with $\tau = 10^4$.

\begin{table}
\centering
\caption{Summary of the iEEG datasets. We report the ID used in Tables \ref{tab:ieeg_corr} and \ref{tab:ieeg_dpli} to identify subjects, the original ID from the datasets, the number of graphs in the nominal and non-nominal distributions, and the number of nodes for each subject.}
\label{tab:ieeg-data}
\begin{tabular}{@{}cccccc@{}}
\toprule
\textbf{Dataset} & \textbf{ID} & \textbf{Original ID} & \textbf{\# graphs ($Q_0$)} & \textbf{\# graphs ($Q_1$)} & \textbf{$N$} \\ \midrule
\multirow{8}{*}{SDC} & D1 & Dog 2 & 1148 & 172 & 16 \\
 & D2 & Dog 3 & 4760 & 480 & 16 \\
 & D3 & Dog 4 & 2790 & 257 & 16 \\
 & D4 & Patient 2 & 2990 & 151 & 16 \\
 & D5 & Patient 5 & 2610 & 135 & 64 \\
 & D6 & Patient 6 & 2772 & 225 & 30 \\
 & D7 & Patient 7 & 3239 & 282 & 36 \\ \midrule
\multirow{3}{*}{SPC} & P1 & Dog 2 & 500 & 42 & 16 \\
 & P2 & Dog 3 & 1440 & 72 & 16 \\
 & P3 & Dog 4 & 804 & 97 & 16 \\ \midrule
\end{tabular}
\end{table}

\subsubsection*{Results}
The proposed method denotes a good performance on iEEG data, where the CCM ensemble, with R-CDT and the geometric discriminator, outperforms single-curvature manifolds and the baseline, on most patients.
In Table \ref{tab:ieeg_corr}, we report the results obtained with Pearson's correlation as functional connectivity measure.
Using DPLI as connectivity measure resulted in a slightly worse performance on average (results are shown in Table \ref{tab:ieeg_dpli}). DPLI is a measure of ``directed'' connectivity, resulting in directed graphs for the functional connectivity networks. As correlation, instead, produces undirected graphs, our results indicate that for this application scenario symmetric connectivity measures might be more suitable in terms of CDT performance. Further connectivity measures will be taken into account in future research.
We notice that the spherical CCM denotes a marginal advantage w.r.t.\ the other configurations on P1, indicating that single CCMs can be effective in some cases. We also notice the poor performance achieved by all configurations on subject P2. Here, when considering preictal graphs, the representation learned by the encoder collapses around the mean value of nominal regime (a phenomenon known as mode collapse), resulting in a poor detection performance. Adding dropout between the ECC layers in the encoder mitigates the issue, but is still not sufficient to achieve the same results obtained for the other patients. 
The benefits of using the geometric discriminator on SDC and SPC are less evident than on synthetic data (due to the graphs having more nodes), but still amount to a significant reduction ($5\%$ on average) in the number of model parameters.

\begin{table*}
\centering
\caption{AUC score on seizure detection, using Pearson's correlation as functional connectivity measure. We report the best results in bold.}
\label{tab:ieeg_corr}
\begin{tabular}{@{}lll|lllllll|lll@{}}
\toprule
& & & \multicolumn{7}{c}{SDC} & \multicolumn{3}{|c}{SDC} \\
\textbf{CCM} & \textbf{CDT} & \textbf{Emb.} & \textbf{D1} & \textbf{D2} & \textbf{D3} & \textbf{D4} & \textbf{D5} & \textbf{D6} & \textbf{D7} & \textbf{P1} & \textbf{P2} & \textbf{P3} \\ \midrule
\multirow{4}{*}{$\M_*$} & \multirow{2}{*}{D-CDT}    & Geom. & 0.99 & 0.99 & 0.99 & 0.93 & 0.99 & 0.66 & 0.99 & 0.22 & 0.19 & 0.87 \\
 &                                                  & Prior & 0.98 & 0.99 & 0.99 & 0.92 & 0.99 & 0.43 & 0.98 & 0.10 & 0.15 & 0.83 \\
 & \multirow{2}{*}{R-CDT}                           & Geom. & \textbf{1.00} & \textbf{1.00} & \textbf{1.00} & 0.94 & \textbf{1.00} & 0.74 & \textbf{1.00} & 0.54 & 0.51 & 0.97 \\
 &                                                  & Prior & \textbf{1.00} & \textbf{1.00} & \textbf{1.00} & \textbf{0.96} & \textbf{1.00} & 0.79 & 0.98 & 0.40 & 0.45 & \textbf{0.98}
\\\midrule
\multirow{4}{*}{$\M_{-1}$} & \multirow{2}{*}{D-CDT} & Geom. & \textbf{1.00} & 0.99 & 0.99 & 0.94 & 0.99 & 0.62 & \textbf{1.00} & 0.64 & 0.25 & 0.90 \\
 &                                                  & Prior & 0.99 & 0.99 & 0.99 & 0.90 & 0.99 & 0.34 & 0.95 & 0.78 & 0.16 & 0.92 \\
 & \multirow{2}{*}{R-CDT}                           & Geom. & 0.99 & 0.99 & 0.99 & 0.93 & 0.98 & 0.85 & 0.99 & 0.25 & 0.28 & 0.90 \\
 &                                                  & Prior & 0.97 & 0.99 & 0.97 & 0.83 & 0.98 & 0.82 & 0.96 & 0.58 & 0.28 & 0.95 \\ \midrule
\multirow{4}{*}{$\M_{0}$} & \multirow{2}{*}{D-CDT}  & Geom. & 0.99 & 0.99 & 0.99 & 0.81 & 0.99 & 0.00 & 0.99 & 0.44 & 0.09 & 0.78 \\
 &                                                  & Prior & 0.99 & 0.99 & 0.99 & 0.74 & 0.97 & 0.59 & 0.99 & 0.76 & 0.10 & 0.85 \\
 & \multirow{2}{*}{R-CDT}                           & Geom. & 0.92 & 0.96 & 0.93 & 0.83 & 0.97 & 0.61 & 0.93 & 0.17 & 0.17 & 0.74 \\
 &                                                  & Prior & 0.91 & 0.93 & 0.88 & 0.62 & 0.97 & 0.71 & 0.96 & 0.15 & 0.53 & 0.65 \\ \midrule
\multirow{4}{*}{$\M_{1}$} & \multirow{2}{*}{D-CDT}  & Geom. & 0.99 & 0.99 & 0.99 & 0.94 & 0.88 & 0.65 & 0.89 & 0.67 & 0.15 & 0.91 \\
 &                                                  & Prior & 0.99 & 0.99 & 0.99 & 0.90 & 0.99 & 0.33 & 0.98 & \textbf{0.87} & 0.48 & 0.82 \\
 & \multirow{2}{*}{R-CDT}                           & Geom. & 0.99 & 0.99 & 0.99 & \textbf{0.96} & 0.99 & 0.85 & 0.97 & 0.57 & 0.20 & 0.95 \\
 &                                                  & Prior & 0.99 & 0.99 & 0.99 & 0.95 & 0.99 & 0.53 & 0.99 & 0.64 & 0.73 & 0.91 \\ \midrule
$\M_0$ & R-CDT & \cite{zambon2017concept}                   & 0.92 & 0.74 & 0.84 & 0.90 & 0.90 & \textbf{0.88} & 0.79 & 0.73 & \textbf{0.84} & 0.90 \\ \bottomrule
\end{tabular}
\end{table*}

\begin{table*}
\centering
\caption{AUC score on seizure detection, using DPLI as functional connectivity measure. We report the best results in bold.}
\label{tab:ieeg_dpli}
\begin{tabular}{@{}lll|lllllll|lll@{}}
\toprule
& & & \multicolumn{7}{c}{SDC} & \multicolumn{3}{|c}{SDC} \\
\textbf{CCM} & \textbf{CDT} & \textbf{Emb.} & \textbf{D1} & \textbf{D2} & \textbf{D3} & \textbf{D4} & \textbf{D5} & \textbf{D6} & \textbf{D7} & \textbf{P1} & \textbf{P2} & \textbf{P3} \\ \midrule
\multirow{4}{*}{$\M_*$} & \multirow{2}{*}{D-CDT} & Geom.     & 0.99 & 0.99 & 0.99 & 0.72 & 0.98 & 0.15 & 0.69 & 0.20 & 0.30 & 0.62 \\
 &  & Prior                                                  & 0.98 & 0.99 & 0.99 & 0.74 & 0.98 & 0.30 & 0.63 & 0.18 & 0.32 & 0.61 \\
 & \multirow{2}{*}{R-CDT} & Geom.                            & \textbf{1.00} & 0.99 & \textbf{1.00} & 0.81 & \textbf{0.99} & 0.68 & 0.81 & 0.63 & 0.61 & 0.73 \\
 &  & Prior                                                  & \textbf{1.00} & \textbf{1.00} & 0.99 & \textbf{0.82} & 0.98 & 0.53 & \textbf{0.85} & 0.55 & 0.65 & 0.73 \\ \midrule
 
\multirow{4}{*}{$\M_{-1}$} & \multirow{2}{*}{D-CDT} & Geom.  & 0.99 & 0.99 & 0.99 & 0.38 & 0.98 & 0.44 & 0.12 & 0.00 & 0.22 & 0.51 \\
 &  & Prior                                                  & 0.99 & 0.99 & 0.99 & 0.51 & 0.98 & 0.18 & 0.30 & 0.00 & 0.21 & 0.63 \\
 & \multirow{2}{*}{R-CDT} & Geom.                            & 0.91 & 0.96 & 0.90 & 0.61 & 0.84 & 0.29 & 0.51 & 0.37 & 0.50 & 0.56 \\
 &  & Prior                                                  & 0.89 & 0.91 & 0.88 & 0.52 & 0.81 & 0.54 & 0.55 & 0.48 & 0.47 & 0.55 \\ \midrule

\multirow{4}{*}{$\M_{0}$} & \multirow{2}{*}{D-CDT} & Geom.   & \textbf{1.00} & \textbf{1.00} & \textbf{1.00} & 0.79 & \textbf{0.99} & 0.34 & 0.84 & 0.30 & 0.37 & 0.57 \\
 &  & Prior                                                  & 0.97 & 0.99 & 0.96 & 0.61 & \textbf{0.99} & 0.70 & 0.32 & 0.35 & 0.37 & 0.66 \\
 & \multirow{2}{*}{R-CDT} & Geom.                            & 0.99 & 0.99 & 0.99 & 0.81 & \textbf{0.99} & 0.51 & 0.84 & 0.24 & 0.49 & 0.63 \\
 &  & Prior                                                  & 0.94 & 0.99 & 0.98 & 0.70 & 0.97 & 0.51 & 0.68 & 0.41 & 0.39 & 0.66 \\ \midrule

\multirow{4}{*}{$\M_{1}$} & \multirow{2}{*}{D-CDT} & Geom.   & 0.99 & 0.99 & 0.99 & 0.70 & 0.94 & 0.38 & 0.72 & 0.31 & 0.50 & 0.48 \\
 &  & Prior                                                  & 0.90 & \textbf{1.00} & 0.97 & 0.58 & 0.96 & 0.41 & 0.41 & 0.41 & 0.52 & 0.50 \\
 & \multirow{2}{*}{R-CDT} & Geom.                            & 0.99 & 0.99 & 0.99 & 0.72 & 0.96 & 0.28 & 0.84 & 0.27 & 0.51 & 0.65 \\
 &  & Prior                                                  & 0.99 & 0.99 & 0.98 & 0.70 & 0.96 & 0.42 & 0.56 & 0.44 & 0.54 & 0.51 \\ \midrule
$\M_0$ & R-CDT & \cite{zambon2017concept}                    & 0.92 & 0.69 & 0.78 & 0.90 & 0.90 & \textbf{0.82} & 0.77 & \textbf{0.73} & \textbf{0.78} & \textbf{0.90} \\ \bottomrule
\end{tabular}
\end{table*}

\subsection{Detection of hostile behaviour} 
\label{sec:action}

As a third application, we consider a practical scenario in computer vision where graphs representing the skeletal structure of human beings are used to perform action recognition. 
In line with our proposed method, the task consists of detecting when the stream of skeletal data changes from a nominal action performed by the subjects, to a non-nominal one. Practical applications of this setting include the surveillance of public places for security reasons, the detection of a distracted driver, or the detection of incidents for people at risk (e.g., children and elderly people). 
For this experiment, we focus on one of such possible tasks, namely on detecting whether the interaction between two subjects changes from friendly to hostile, using skeletal data extracted from video samples. 
Because skeletal data provides information and constraints on the overall pose of subjects that are not explicitly encoded in raw video, approaches based on graph neural networks have been successfully applied to achieve state-of-the-art results in action recognition \cite{stgcn2018aaai}, denoting a better performance than traditional deep learning algorithms.

For this experiment we consider the NTU RGB+D dataset for action recognition \cite{Shahroudy_2016_CVPR}, a large collection of video samples containing RGB images, infrared depth maps, and skeletal data of 56880 action samples. 
The dataset contains 60 different action types, including daily, mutual, and health-related actions. Actions are performed by 40 volunteers and each action is repeated twice in 17 different camera settings. The dataset consists of short clips ($\sim 2$ seconds long) sampled at 30Hz. Skeletal data are provided for each frame of the clips. Each subject is represented by 25 joints annotated with 3D coordinates and orientation (i.e., position, direction and rotation of the joint in space), and 2D position w.r.t.\ the frame for both the RGB and infrared channels. Metadata regarding the confidence of the measurement (i.e., whether the annotation is missing, inferred, or actually recorded) is also provided. The topological connections between pairs of joints are fixed and known \textit{a priori}.

\begin{figure}
    \centering
    \includegraphics[width=\linewidth]{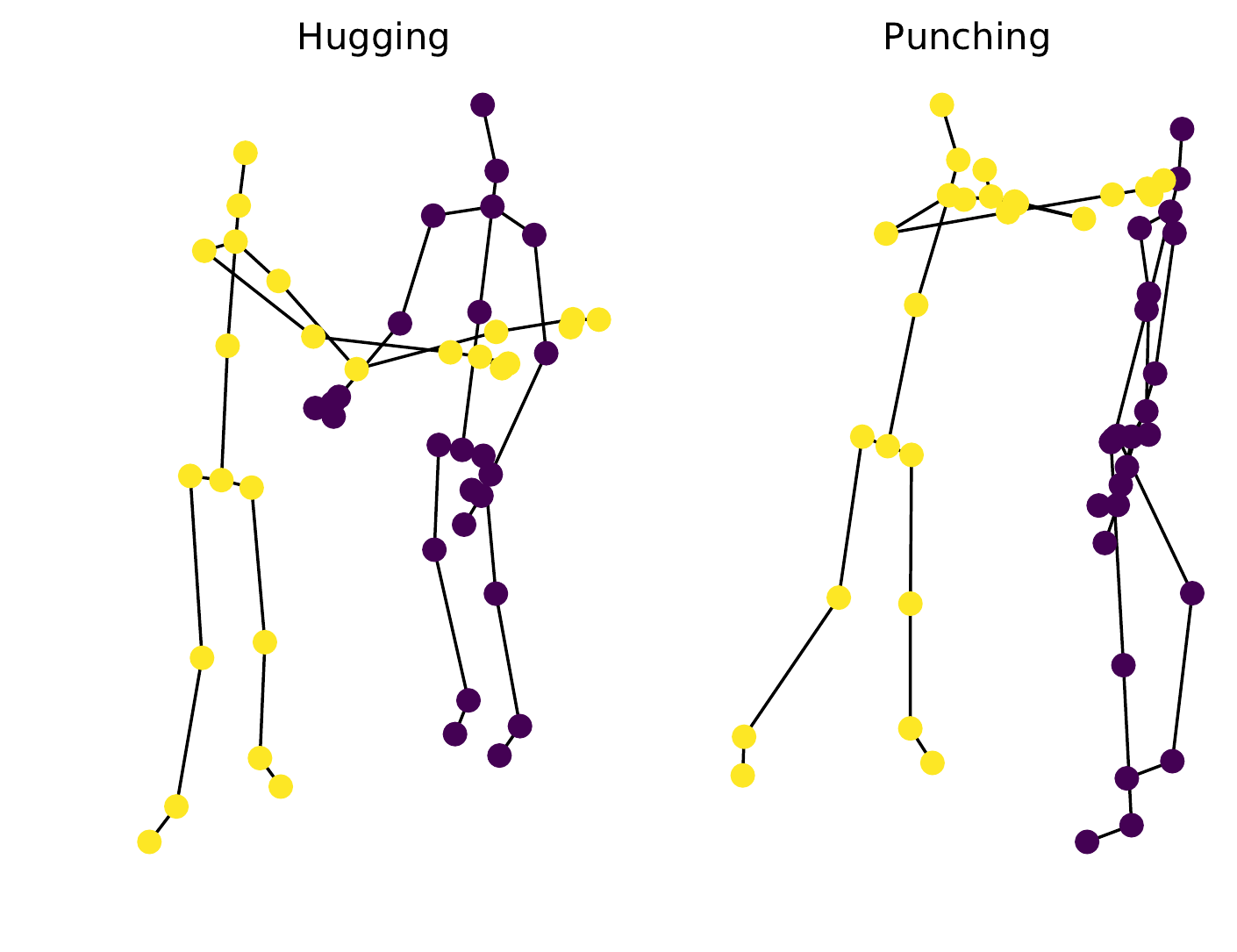}
    \caption{Examples of graphs sampled from NTU RGB+D, respectively from \textit{hugging} and \textit{punching} action classes. The nodes associated to a specific individual are colour-coded accordingly, although the AAE is fed with the disjoint union of the graphs without additional information. Best viewed in colour.}
    \label{fig:actions}
\end{figure}

To test our methodology, we consider a subset of NTU RGB+D containing mutual interactions, namely the \textit{hugging} and \textit{punching} actions (examples shown in Figure \ref{fig:actions}), where each skeletal sample consists of the disjoint union of the two graphs representing the interacting subjects. 
The task is then defined as detecting when the interaction between the two subjects changes from friendly (\textit{hugging}) to hostile (\textit{punching}).

The graph stream is generated in the same fashion as in the previous two experiments, by sampling graphs from the \textit{hugging} distribution for the nominal regime, and then switching to the \textit{punching} distribution to simulate a change. 
The assumption of stationarity of $Q_0$ required to configure and run the CDT, which was met for the previous two experiments\footnote{The Delaunay triangulations are i.i.d.\ by construction, while the functional connectivity networks are extracted from independent clips.}, here cannot be met due to each individual clip showing a high correlation between successive samples, which is reflected also on the embeddings. To avoid this issue, one possibility is to lower the sampling rate with which the skeletal data is acquired, in order to decorrelate the observations. However, we note that lowering the sampling rate in order to obtain an i.i.d.\ graph stream is equivalent, in this controlled setting, to taking random permutations of the available data. This results in a stationary stream, and allows us to test our method without having to purposefully waste precious samples.
The training data is thus obtained by randomly sampling graphs from the \textit{hugging} distribution. Similarly, for the operational test stream, we first randomly sample graphs from \textit{hugging}, and then change to \textit{punching}. While in principle it is not necessary to randomise the non-nominal regime, we keep the same setting to ensure that the CDT only detects changes in the actual graph class, rather than in the sampling technique. 
To have independence between the training and operational streams, we sample the former from the first set of clips recorded for each subject, while the latter is sampled from the second set of repetitions. 
The number of graphs for each regime in the training and operational streams is reported in Table \ref{tab:action-data}, and correspond to the number of graphs from the respective sets of repetitions.

\begin{table}
    \centering
    \caption{Number of graphs sampled for the training and operational phases. Training graphs correspond to the first clips recorded for each subject, while operational graphs are taken from the second repetitions.}
    \label{tab:action-data}
    \begin{tabular}{lcc}
        \toprule
        \textbf{Phase} & \textbf{Action}  & \textbf{\# of graphs} \\
        \midrule
        Train & Hugging  & 26818 \\
        \midrule
        \multirow{2}{*}{Operational} & Hugging  & 26166 \\
        & Punching & 24512 \\
        \bottomrule
    \end{tabular}
\end{table}

We leverage the results obtained in the previous two experiments for configuring the change detection pipeline. Accordingly, we report results obtained with the ensemble of manifolds $\mc M_*$, using the geometric discriminator and R-CDT.

\subsubsection*{Results}
\begin{table}
\centering
\caption{Performance of R-CDT and D-CDT on detection of hostile behaviour.}
\label{tab:action}
\begin{tabular}{@{}lll|cc@{}}
\toprule
\textbf{CCM} & \textbf{Emb.} & \textbf{Emb.} & AUC & ARL \\ 
\midrule
\multirow{2}{*}{$\M_*$} & \multirow{2}{*}{Geom} & R-CDT & \textbf{0.999} & \textbf{1.04} \\
 & & D-CDT & 0.965 & 3.27 \\
\bottomrule
\end{tabular}
\end{table}

Running R-CDT on the proposed graph stream results in an AUC score of $0.999$, i.e., the algorithm is consistently able to identify changes in stationarity with a very short delay. The average run length (ARL) obtained by R-CDT in the non-nominal regime is $\sim 1.04$, meaning that an alarm is raised by the algorithm almost at every window. By comparison, D-CDT denotes a similarly good separability in the distributions of the run lengths, with AUC $0.965$ (see Table \ref{tab:action}). However, the ARL of D-CDT is significantly higher at $3.27$, indicating a slower detection of the change. 

\begin{figure}
    \centering
    \includegraphics[width=\linewidth]{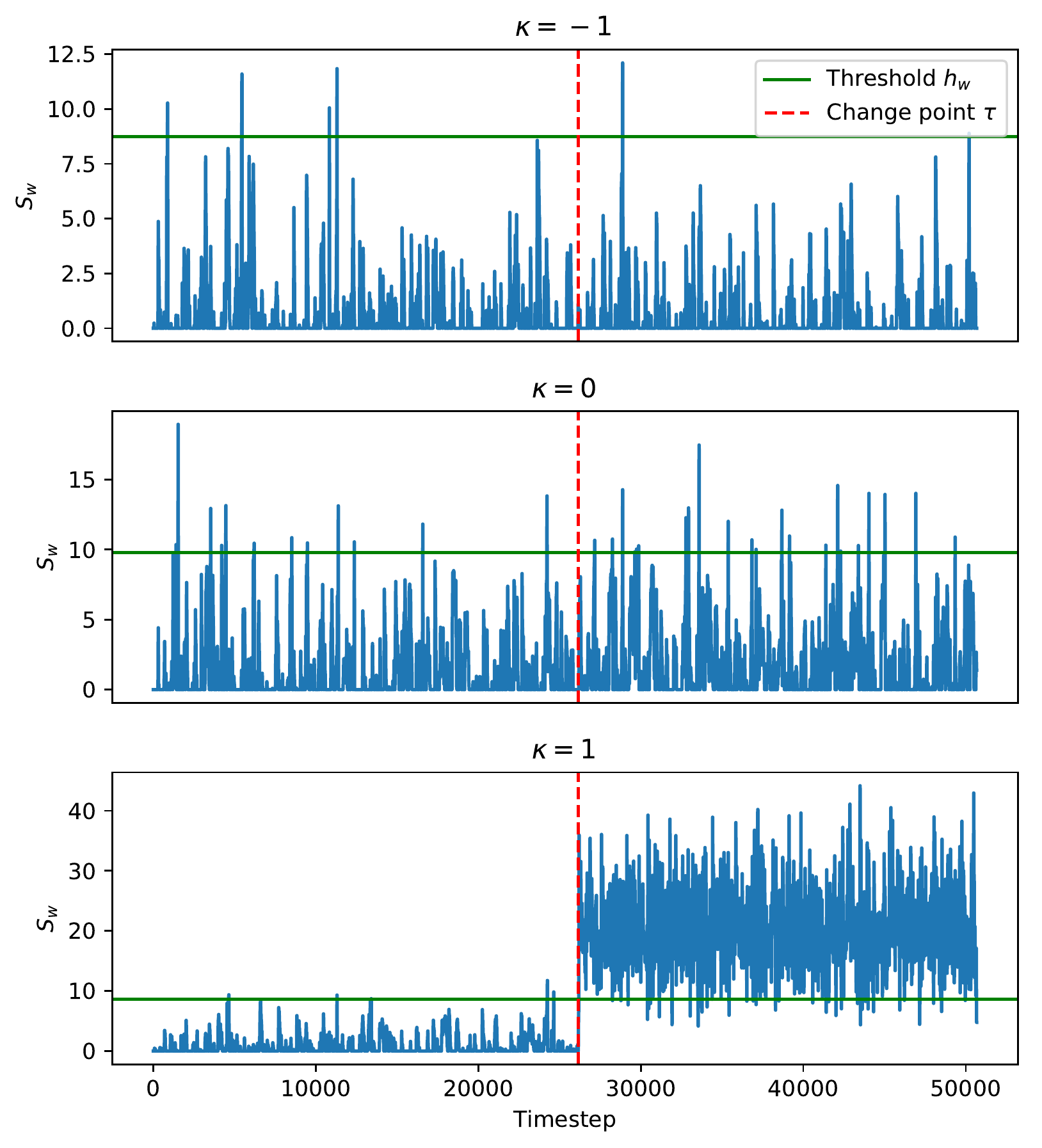}
    \caption{Accumulator $S_w$ on the operational stream, for the three independent CDTs run by R-CDT on the ensemble of manifolds. The dashed red line denotes the change point of the stream. Whenever the accumulator exceeds the threshold (green line) an alarm is raised. The CDT performed on the spherical manifold ($\kappa=1$) is able to identify the change in stationarity, while the Euclidean and hyperbolic embeddings do not show such a strong response to the change.}
    \label{fig:rcdt_action}
\end{figure}

\begin{figure*}
    \centering
    \includegraphics[width=0.8\linewidth]{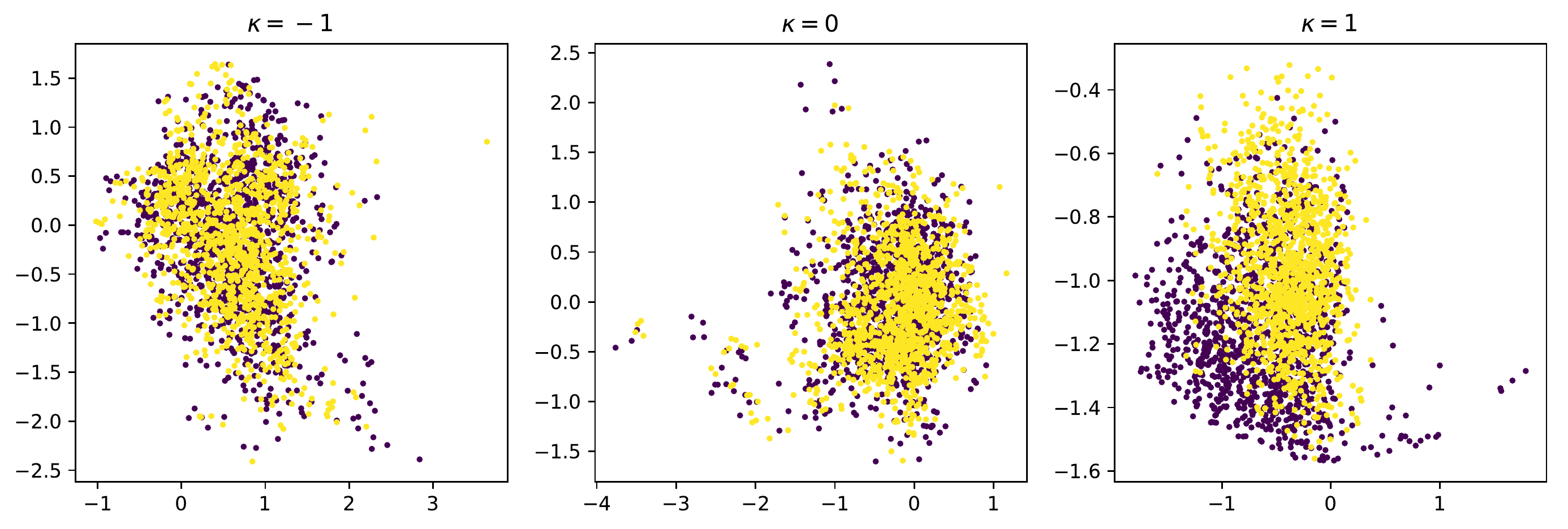}
    \caption{Embeddings learned by the AAE with latent ensemble of CCMs, $\M_*$. We show a planar projection for the hyperbolic and Euclidean CCMs, and the Hammer projection of the spherical CCM, for the two regimes (\textit{hugging} is in yellow, \textit{punching} in purple). Best viewed in colour.}
    \label{fig:embeddings_action}
\end{figure*}

In Figure \ref{fig:rcdt_action}, we show the evolution of the accumulator $S_w$ on the operational stream for each of the three CDTs run by R-CDT (one for each manifold).
It is possible to note how the spherical component of the embeddings computed by the AAE provides clear indication of the existence of a change in stationarity. A similar conclusion is also evident by comparing the distribution on the spherical embeddings to the other two geometries (shown in Figure \ref{fig:embeddings_action}).

Results highlight the advantages of using an ensemble of CCMs for learning a representation and stress the importance of representing graph-structured data by using non-Euclidean domains. 

\section{Conclusion}
\label{sec:conclusion}

In this paper, we introduced a novel data-driven method for detecting changes in stationarity in a stream of attributed graphs.
The methodology is based on an adversarial autoencoder that embeds graphs on constant-curvature manifolds, onto which we apply statistical and geometrical tools for the analysis.
Experimental results demonstrated the effectiveness of what proposed by considering streams of graph-structured data for both synthetic and real-world applications.
Our results showed that the ensemble of CCMs ($\M_*$), the geometric discriminator, and the Riemannian version of the CDT consistently yield the best detection performance, making this configuration a safe choice when no prior information is available about the problem at hand.
We believe that the proposed framework can be easily extended beyond the scope considered in this paper, as many application domains are characterised by graphs that change over time, such as in sensor, wireless, and gene expression networks.

\subsection*{Acknowledgements}
This research is funded by the Swiss National Science Foundation project 200021\_172671: ``ALPSFORT: A Learning graPh-baSed framework FOr cybeR-physical sysTems''. We gratefully acknowledge the support of NVIDIA Corporation with the donation of the Titan Xp GPU used for this research.
LL gratefully acknowledges partial support of the Canada Research Chairs program.

\bibliography{main}
\bibliographystyle{IEEEtran}

\newpage
\begin{IEEEbiography}[{\includegraphics[scale=0.175,keepaspectratio]{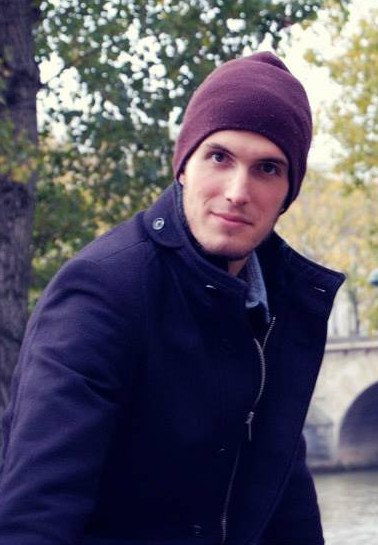}}]{Daniele Grattarola}
rececived his M.Sc. degree \textit{magna cum laude} in Computer Science and Engineering from Politecnico di Milano (Italy), in 2017. He is a Ph.D.\ student with the Faculty of Informatics at Università della Svizzera italiana, in Lugano (Switzerland). His research interests include graph neural networks, graph stochastic processes, epilepsy, and reinforcement learning. 
\end{IEEEbiography}
\vskip -2\baselineskip plus -1fil
\begin{IEEEbiography}[{\includegraphics[scale=0.15,keepaspectratio]{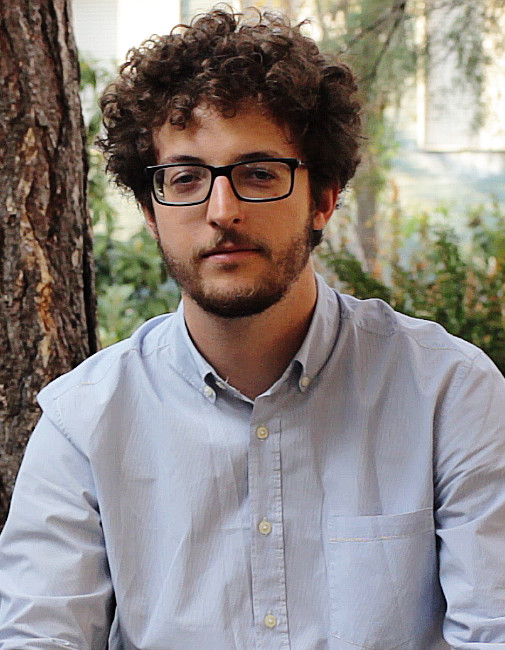}}]{Daniele Zambon}
received the M.Sc. degree in Mathematics from Universit\`a degli Studi di Milano, Italy, in 2016. 
Currently he is Ph.D. student with Faculty of Informatics, Universit\`a della Svizzera italiana, Lugano, Switzerland.
His research interests include: graph representation, statistical processing of graph streams, change and anomaly detection.
\end{IEEEbiography}
\vskip -2\baselineskip plus -1fil
\begin{IEEEbiography}[{\includegraphics[scale=0.3,keepaspectratio]{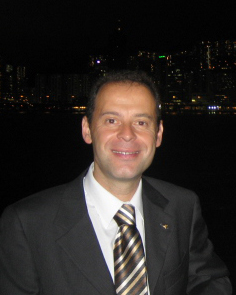}}]{Cesare Alippi}
(F'06) received the degree in electronic engineering cum laude in 1990 and the PhD in 1995 from Politecnico di Milano, Italy. Currently, he is a Full Professor of information processing systems with Politecnico di Milano, Italy, and of Cyber-Phsical and embedded systems at the Universita' della Svizzera Italiana, Switzerland. He has been a visiting researcher at UCL (UK), MIT (USA), ESPCI (F), CASIA (RC), A*STAR (SIN).
Alippi is an IEEE Fellow, Distinguished lecturer of the IEEE CIS, Member of the Board of Governors of INNS, Vice-President education of IEEE CIS, Associate editor (AE) of the IEEE Computational Intelligence Magazine, past AE of the IEEE-Trans. Instrumentation and Measurements, IEEE-Trans. Neural Networks, and member and chair of other IEEE committees. 
In 2004 he received the IEEE Instrumentation and Measurement Society Young Engineer Award; in 2013 he received the IBM Faculty Award. He was also awarded the 2016 IEEE TNNLS outstanding paper award and the 2016 INNS Gabor award.
Among the others, Alippi was General chair of the International Joint Conference on Neural Networks (IJCNN) in 2012, Program chair in 2014, Co-Chair in 2011. He was General chair of the IEEE Symposium Series on Computational Intelligence 2014. 
Current research activity addresses adaptation and learning in non-stationary environments and Intelligence for embedded systems. 
Alippi holds 5 patents, has published in 2014 a monograph with Springer on ``Intelligence for embedded systems'' and (co)-authored more than 200 papers in international journals and conference proceedings.
\end{IEEEbiography}
\vskip -2\baselineskip plus -1fil
\begin{IEEEbiography}[{\includegraphics[scale=0.026,keepaspectratio]{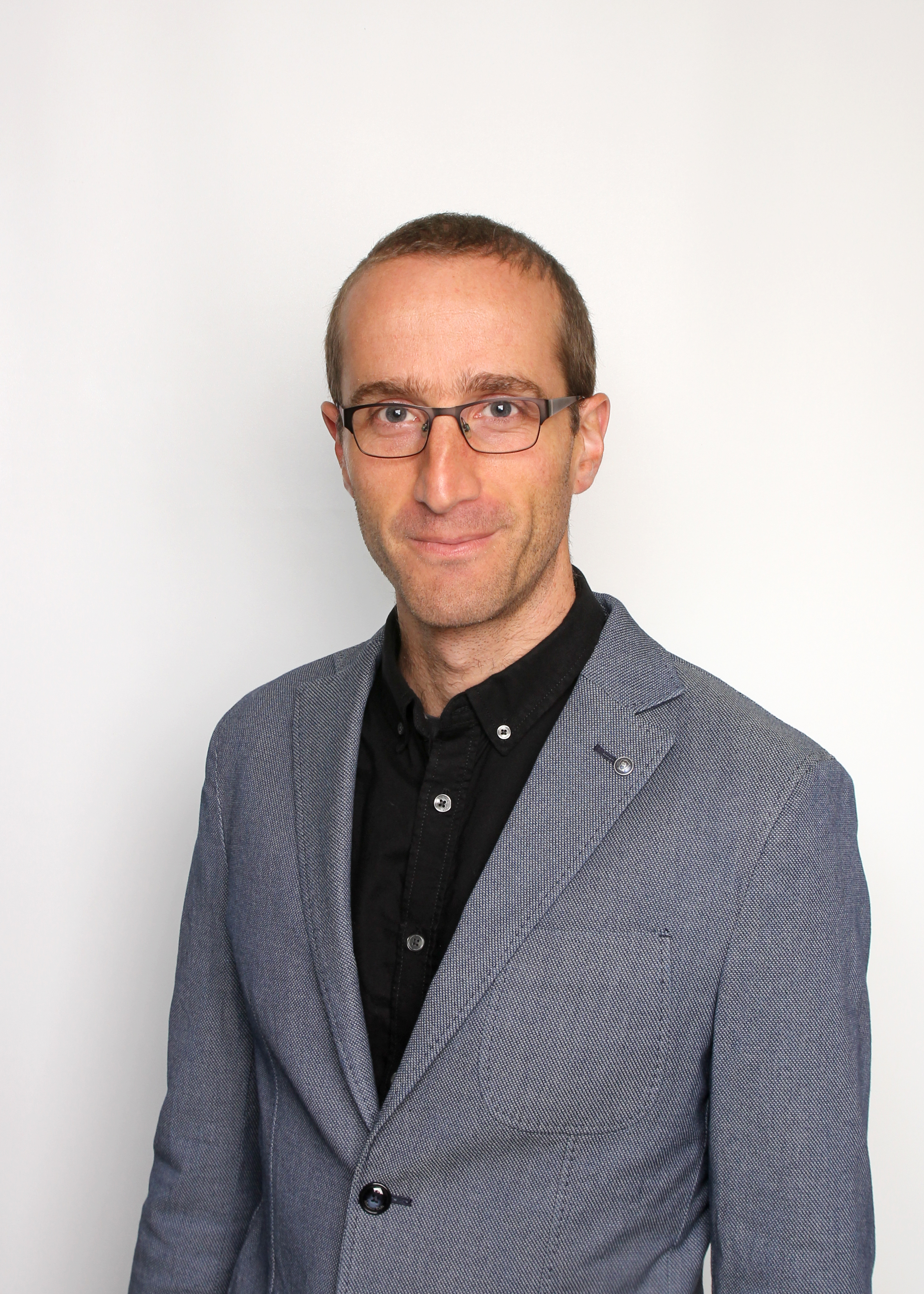}}]{Lorenzo Livi} (M'14) received the B.Sc. degree and M.Sc. degree from the Department of Computer Science, Sapienza University of Rome, Italy, in 2007 and 2010, respectively, and the Ph.D. degree from the Department of Information Engineering, Electronics, and Telecommunications at Sapienza University of Rome, in 2014. He has been with the ICT industry during his studies. From January 2014 until April 2016, he was a Post Doctoral Fellow at Ryerson University, Toronto, Canada. From May 2016 until September 2016, he was a Post Doctoral Fellow at the Politecnico di Milano, Italy and Universita' della Svizzera Italiana, Lugano, Switzerland.
Currently, he is an Assistant Professor jointly appointed with the Departments of Computer Science and Mathematics at the University of Manitoba, Canada. He is also a Lecturer (Assistant Professor) in Data Science at the Department of Computer Science, University of Exeter, United Kingdom. In November 2018, Dr. Livi was awarded the prestigious Tier 2 Canada Research Chair in Complex Data. He is an Associate Editor of the IEEE-TNNLS, Applied Soft Computing (Elsevier) and a regular reviewer for several international journals, including IEEE Transactions and Elsevier journals. His research interests include Machine Learning, Time Series Analysis and Complex Dynamical Systems, with focused applications in Systems Biology and Computational Neuroscience.
\end{IEEEbiography}
\end{document}